\documentclass[10pt,twocolumn,letterpaper]{article}

\usepackage[pagenumbers]{wacv} 
\usepackage{booktabs}
\usepackage{multirow}
\usepackage{graphicx}
\usepackage{caption}

\usepackage{pifont}
\usepackage{array}
\usepackage{arydshln}

\newcommand{\cmark}{\ding{51}}
\newcommand{\xmark}{\ding{55}}

\definecolor{wacvblue}{rgb}{0.21,0.49,0.74}
\usepackage[pagebackref,breaklinks,colorlinks,allcolors=wacvblue,hypertexnames=false]{hyperref}

\def\wacvPaperID{1376} 
\def\confName{WACV}
\def\confYear{2027}

\title{Decoupled Guidance: Disentangling Subject and Context Pathways in Text-to-Image Personalization}

\author{
Seongmin Kim\textsuperscript{*,1}\quad
Kyucheol Shin\textsuperscript{*,1}\quad
Heesun Jung\textsuperscript{2}\quad
Jinseo Kim\textsuperscript{1}\quad
Sungyong Baik\textsuperscript{1,2\dag}\\[0.5em]
\textsuperscript{1}Department of Artificial Intelligence,
\textsuperscript{2}Department of Data Science\\
Hanyang University \\
{\tt\small \{smin28, akami40, jheesun, jinseo84, dsybaik\}@hanyang.ac.kr}\\[0.3em]
\textsuperscript{*}Equal contribution\quad
\textsuperscript{\dag}Corresponding author
}

\newcommand{\mymethod}{DeGu}

\usepackage{amsmath,amsfonts,bm}

\def\eqref#1{equation~\ref{#1}}

\def\1{\bm{1}}

\def\vc{{\bm{c}}}

\DeclareMathAlphabet{\mathsfit}{\encodingdefault}{\sfdefault}{m}{sl}
\SetMathAlphabet{\mathsfit}{bold}{\encodingdefault}{\sfdefault}{bx}{n}

\begin{document}

\twocolumn[{%
    \maketitle
    \begin{center}
    \includegraphics[width=0.85\linewidth]{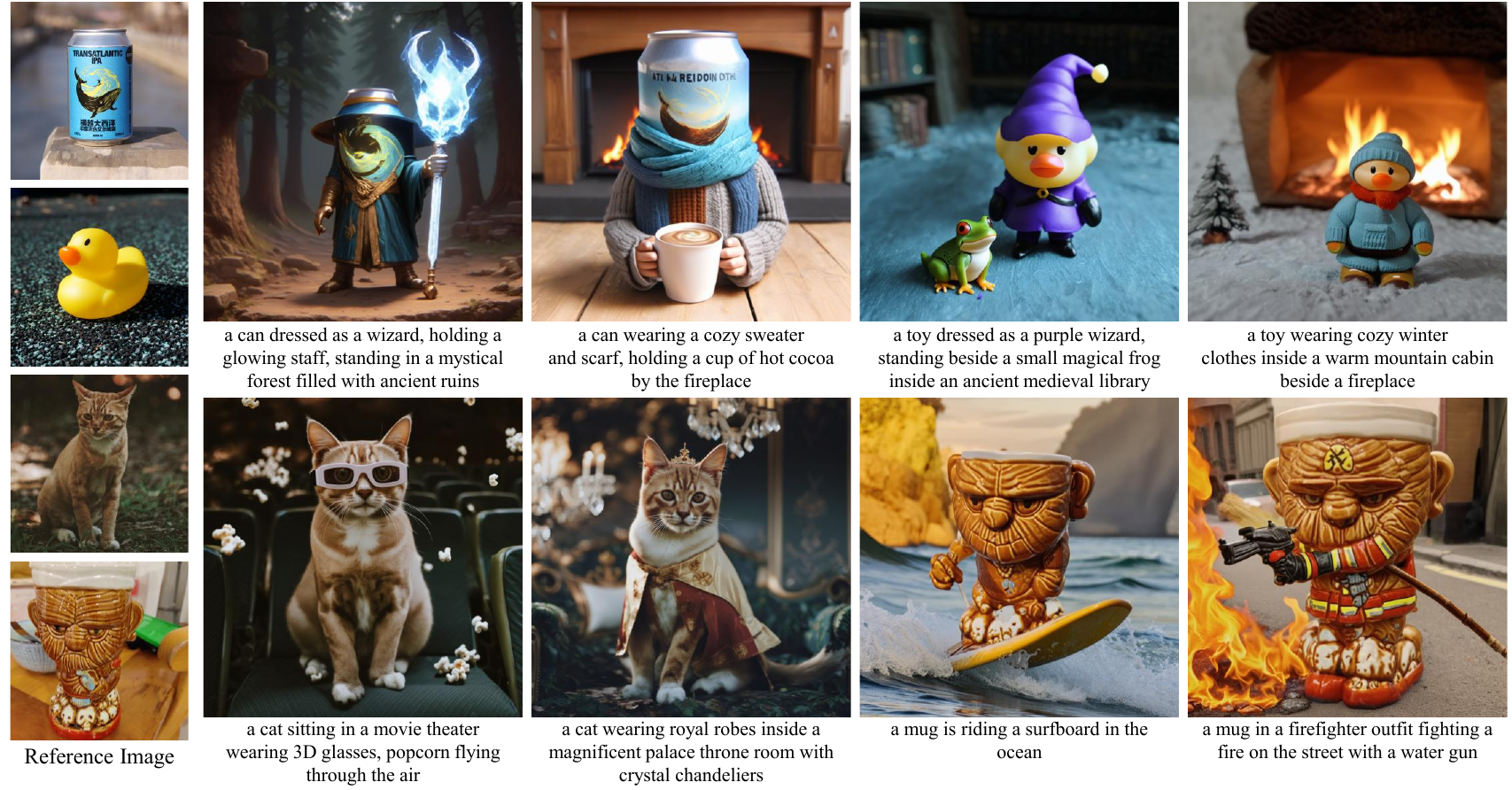}
        \captionof{figure}{\textbf{Personalization results from DeGu.} Given a few reference images, our method faithfully preserves fine-grained subject identity while composing each subject into novel scenes. All images are generated with the SD v3.5 backbone.}
        \label{fig:results_summary}
    \end{center}
    \vspace{1em}
}]

\begin{figure}[t]
    \centering
    \includegraphics[width=\linewidth]{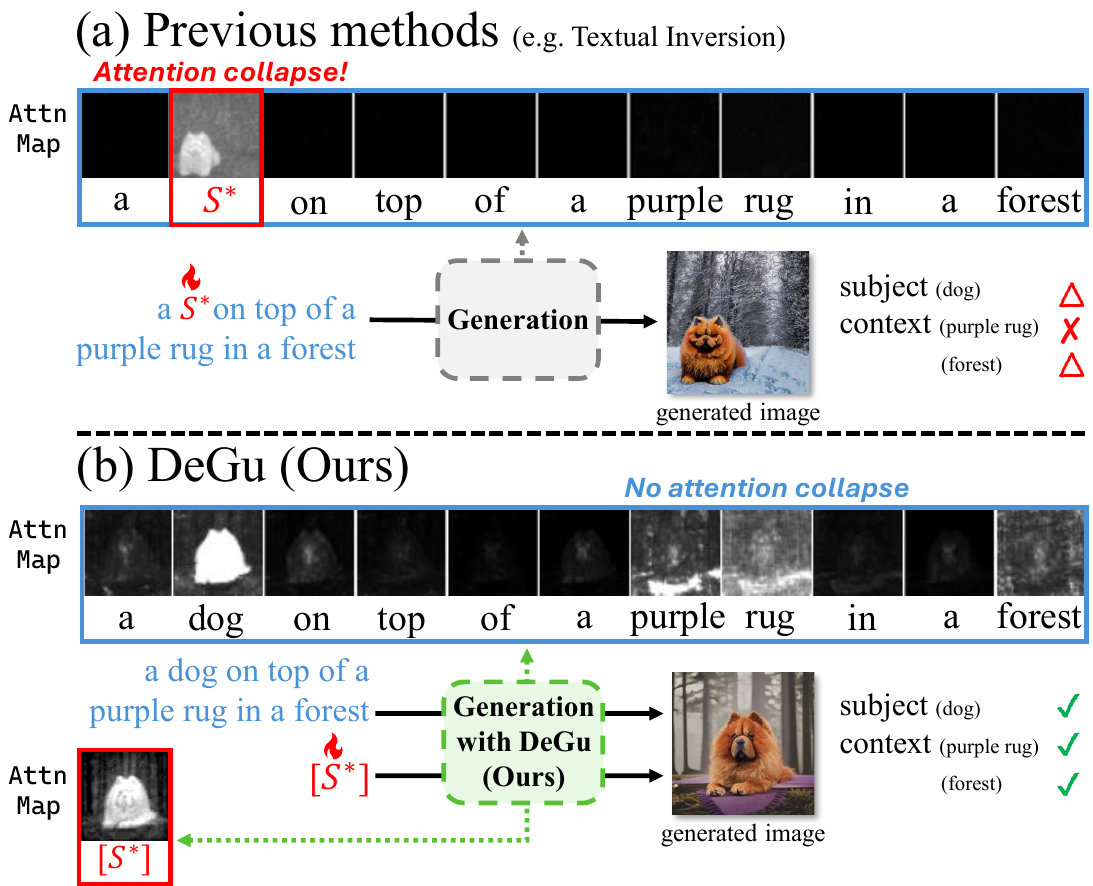}
    \caption{\textbf{Conditioning entanglement in existing personalization and the proposed \mymethod{} framework.}
    (a) Previous methods route subject identity and scene context through a single conditioning pathway.
    The attention maps illustrate attention collapse in embedding optimization (detailed analysis in Section~\ref{sec:analysis}).
    (b) \mymethod{} routes subject identity and scene context through two independent guidance streams, preventing the attention competition between subject and context.
    }
    \label{fig:teaser}
\end{figure}

\begin{abstract}

Text-to-image personalization aims to generate a user-provided subject in novel scenes described by text.
However, most existing methods encode subject identity (fidelity) and context (editability) through the same conditioning pathway, forcing the two to compete for attention-map resources.
We refer to this phenomenon as conditioning entanglement and show that it induces a fidelity-editability trade-off.
We further provide causal evidence by replacing the target subject token with a generic subject token, which produces shifts in attention allocation and corresponding changes in context adherence.
To this end, we propose \textit{Decoupled Guidance} (\mymethod{}), a plug-and-play framework that routes subject identity and scene context through two independent guidance streams.
We further introduce a spatial mixing mechanism that dynamically fuses these streams, ensuring each operates within its semantically relevant region without interference.
Furthermore, \mymethod{} can be readily applied to existing personalization methods without modifying the underlying backbone models, consistently improving the overall personalization performance while enabling inference-time control over the fidelity--editability balance, across diverse methods and backbones, including flow-matching Diffusion Transformers (DiTs). 

\end{abstract}

\section{Introduction}

Recent text-to-image models~\cite{ddpm, ldm, imagen, dalle2, glide, sdxl} can generate photorealistic images from text descriptions such as ``a dog in a café.''
However, these models produce generic subjects rather than specific ones---e.g., unable to generate images of a particular pet with distinctive markings.
Text-to-image personalization~\cite{dreambooth,textualinversion,ms-diffusion} aims to address this limitation: given only a handful of reference images of a target subject (e.g., a pet, a product, or a personal object), the goal is to synthesize the target subject in novel scenes while preserving its identity.
This capability has broad practical values, spanning creative content generation, e-commerce, and personal media production.

However, personalization is a challenging problem that requires a model to simultaneously satisfy two objectives: reproducing a target subject faithfully and following the scene, layout, and attributes described by text.
One promising approach centers on fine-tuning methods~\cite{dreambooth,customdiffusion,disenbooth,svdiff}, which directly adapt model parameters to learn new subjects.
Despite high fidelity, directly adapting model parameters embeds subject identity into the model weights, which limits the extent to which inference-time guidance can steer scene composition. 
On the other hand, \textit{embedding optimization}~\cite{textualinversion,p+,neti,crossinit,dciico} keeps the model frozen and learns a small set of token embeddings representing the new subject.
While this design avoids overwriting prior knowledge, higher subject fidelity tends to come at the cost of text-image alignment.
Meanwhile, encoder-based methods~\cite{elite, ms-diffusion, BLIP-diffusion, SSR} predict subject representations directly from images, eliminating per-subject optimization, but still lack a principled mechanism to independently adjust the balance between subject fidelity and context adherence at inference time.
Despite their differences, these paradigms share a common structural property: subject identity and scene context are routed through a single conditioning pathway, limiting independent control over each objective.

\begin{figure}[t]
 \centering
 \includegraphics[width=0.95\linewidth]{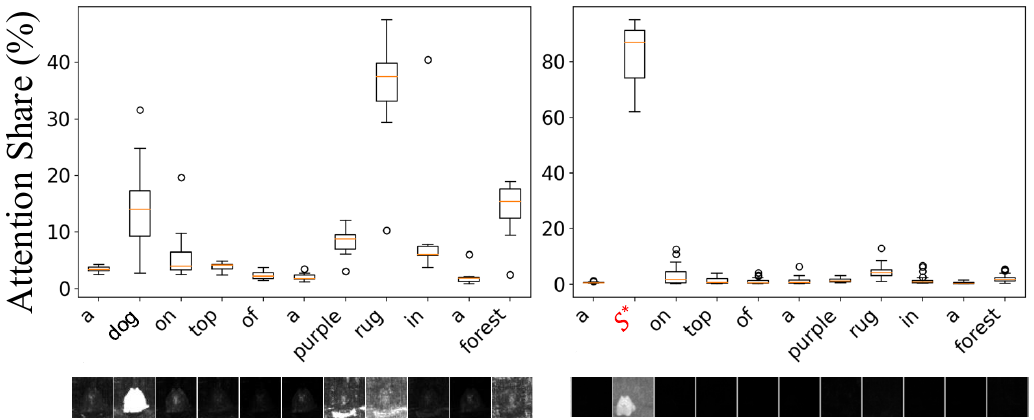}
    \caption{\textbf{Attention collapse is a systematic consequence of unified prompts.}
        Per-token cross-attention magnitudes for a compositional prompt.
        \textbf{Left:} In the pre-trained model, attention distributes evenly across tokens.
        \textbf{Right:} After personalization via TI, the learned subject token $S^*$ absorbs most of the cross-attention budget, suppressing context tokens such as ``purple,'' ``rug,'' and ``forest'' to near zero---an effect we refer to as \textit{attention collapse} (analyzed in Section~\ref{sec:analysis}).
        }
 \label{fig:attention_analysis}
\end{figure}

We examine this shared-pathway structure in the embedding optimization setting, where its effects are most directly observable.
In Section~\ref{sec:analysis}, we show that unified prompts cause the learned subject token to absorb the majority of cross-attention resources---a phenomenon we term \textit{attention collapse}---systematically suppressing context tokens and degrading editability.
Motivated by this analysis, we propose \textbf{Decoupled Guidance (\mymethod{})}, a plug-and-play framework that routes subject identity and scene context through two independent guidance streams, as shown in Figure~\ref{fig:teaser}.

While \mymethod{} is motivated by the attention collapse diagnosed in embedding optimization, the decoupling mechanism itself does not depend on any specific paradigm or backbone architecture.
Experiments show that \mymethod{} consistently improves personalization performance across embedding optimization, fine-tuning, and encoder-based methods on multiple backbones including flow-matching Diffusion Transformers (DiTs), while enabling inference-time control over the fidelity--editability balance through independent guidance scales for each stream.

\begin{figure*}[t!]
    \centering
    \includegraphics[width=\linewidth]{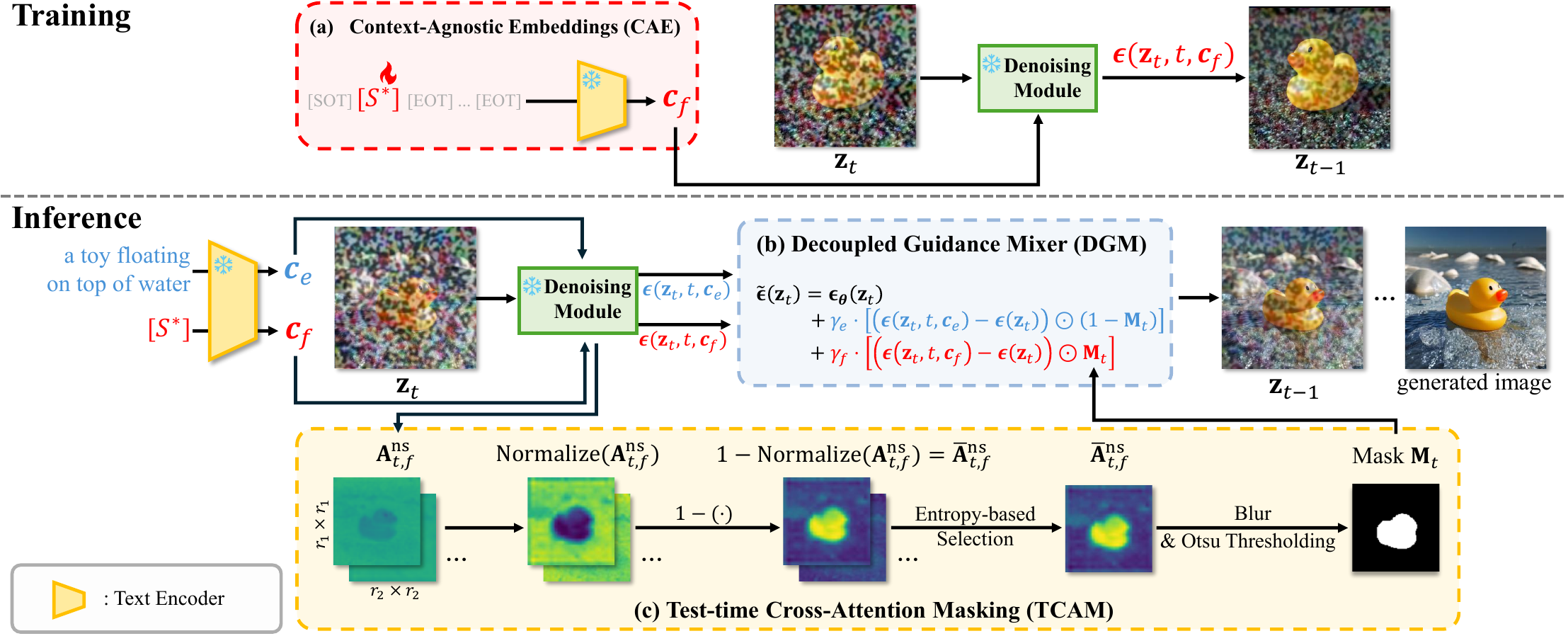}
    \caption{
    \textbf{Overview of our proposed framework.}
    \textcolor{black}{(a) We isolate the subject token from the context tokens and introduce \textit{Context-Agnostic Embeddings} (CAE), which trains the token embedding of our learnable token $[S^\ast]$ using the token sequence $[\text{SOT}]\ [S^\ast]\ [\text{EOT}] \cdots [\text{EOT}]$ to obtain the fidelity condition $\vc_f$ via text encoder $\mathcal{E}$ to focus on capturing the subject identity.}
    (b) At inference, our \textit{Decoupled Guidance Mixer} (DGM) combines $\vc_f$ (for fidelity) with the context condition $\vc_e$ (for editability).
    (c) \textit{Test-time Cross-Attention Masking} (TCAM) derives a subject mask $\mathbf{M}_t$ by inverting and binarizing the attention map of a designated non-subject token in the $\vc_f$ stream.
    This mask spatially partitions the guidance, applying $\vc_f$ inside the subject region and $\vc_e$ to the background (details in Appendix~\ref{sec:supp_implementation}).
    }
    \label{fig:overview}
\end{figure*}
\section{\textcolor{black}{Related work}}\label{sec:relatedwork}

\noindent\textbf{Text-to-image diffusion models.}
Diffusion models~\cite{diffusion,ddpm,ddim,scorebased,flowmatching,zerosnr,hang2024improved}, particularly Latent Diffusion Models (LDM)~\cite{ldm,sdxl} and their popular instantiation Stable Diffusion (SD), have become a widely adopted framework for text-to-image generation, owing to their text-aligned synthesis quality.
Their generation quality stems from large-scale training~\cite{dalle2,imagen,sdxl,glide}, while text alignment is achieved by employing pretrained text encoders such as CLIP~\cite{clip}, BLIP~\cite{blip}, and T5~\cite{t5}, together with Classifier-Free Guidance (CFG)~\cite{cfg}.
Notably, attention layers in these models have been shown to encode spatially meaningful representations useful for segmentation~\cite{diffseg,seediff}.
Despite this progress, generating images of specific unseen subjects remains an open challenge, which we address in this work.

\noindent\textbf{Text-to-image personalization.}
The core challenge of personalization is to faithfully reproduce a specific subject while composing it into novel scenes described by text.
Existing approaches fall broadly into three paradigms.
Embedding optimization methods~\cite{textualinversion,p+,neti,crossinit,conceptdecomposition,hybridbooth,cusconcept,dciico,prospect} keep the model frozen and embed subject characteristics into learnable token embeddings, preserving prior knowledge with minimal storage, but the limited capacity of token embeddings makes it difficult to achieve both high subject fidelity and strong editability~\cite{perfusion, elite}.
Fine-tuning methods~\cite{dreambooth,customdiffusion,svdiff,lora,perfusion,hyperdreambooth,disenbooth,10943814,vico,mixofshow,breakascene,cones,cones2} directly adapt model parameters to learn subject characteristics, achieving high fidelity at the risk of catastrophic forgetting~\cite{overcommingcatastrophic,catastrophicforgetting,learningwithoutforgetting}.
Encoder-based methods~\cite{elite, ms-diffusion, BLIP-diffusion, SSR} train an image encoder to predict the subject directly from reference images, removing per-subject optimization but requiring large-scale datasets for encoder training.
Our method belongs to the embedding optimization paradigm but mitigates the fidelity--editability trade-off by decoupling subject and context conditioning, with only a few hundred kilobytes of storage. It also serves as a plug-and-play module that consistently improves existing methods without modifying their weights.

\noindent\textbf{Attention manipulation in compositional generation.}
A separate line of work~\cite{attendandexcite, prompttoprompt} reweights or redistributes cross-attention during inference to mitigate token neglect in compositional generation.
While these methods offer some improvements, they still operate within a single unified prompt and fail to decouple the conditioning pathways.
Composable diffusion~\cite{Liu2022ComposableDiffusion} combines independent text conditions at the score level via a product-of-experts formulation; our method shares this score-level combination but addresses a different setting where one condition is a learned subject embedding, requiring a dedicated training strategy to establish the conditional independence that text-only composition assumes by construction (Section~\ref{sec:dgm}).
Despite these varied efforts, a structural property persists across personalization methodologies: subject and context conditions share the same attention budget.

\section{The anatomy of conditioning entanglement}
\label{sec:analysis}

In this section, we aim to analyze conditioning entanglement and how it affects personalization.
Prior work attributes the fidelity--editability trade-off to limited token capacity and proposes to expand the conditioning budget~\cite{p+,neti,dciico}.
In contrast, we argue that unified prompts inherently cause \textit{attention competition}, with severe cases yielding \textit{attention collapse} where subject tokens absorb most attention resources.
We test this through three experiments that progressively rule out alternative explanations.

\noindent\textbf{Attention competition in unified prompts.}
All experiments in this section are conducted on SD v2.1 across 39 subjects from diverse categories, each evaluated on 25 compositional prompts.
To determine whether unified prompts systematically bias attention toward the subject token, we analyze cross-attention distributions in the pre-trained model and Textual Inversion~\cite{textualinversion} (TI)---which represents the subject as a learned token $S^*$ in the prompt.
We compute per-token cross-attention magnitude averaged over spatial locations and diffusion timesteps.
For each prompt, we define the \textit{Attention Collapse Score} (ACS) as the fraction of total cross-attention captured by the subject token:
\begin{equation}
    \mathrm{ACS} = \frac{\bar{A}_{S^*}}{\sum_i \bar{A}_{w_i}} \times 100\%,
    \label{eq:acs}
\end{equation}
where $\bar{A}_{w_i}$ denotes the mean cross-attention magnitude of token $w_i$.

\begin{figure}[t]
    \centering
    \begin{subfigure}[t]{0.49\linewidth}
        \centering
        \caption{}
        \label{fig:attention_editability_correlation}
        \includegraphics[width=\linewidth]{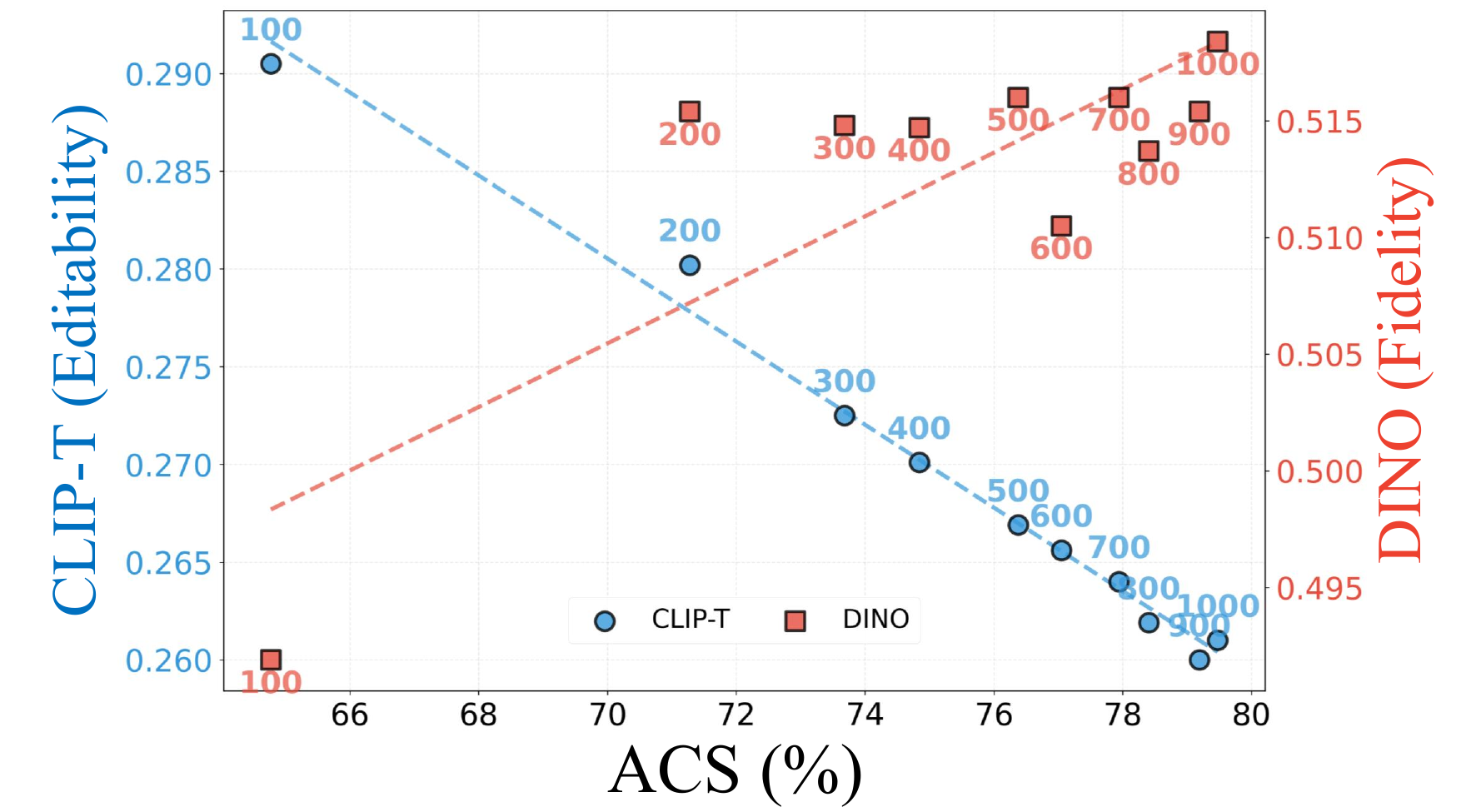}
    \end{subfigure}
    \hfill
    \begin{subfigure}[t]{0.49\linewidth}
        \centering
        \caption{}
        \label{fig:capacity_tradeoff}
        \includegraphics[width=\linewidth]{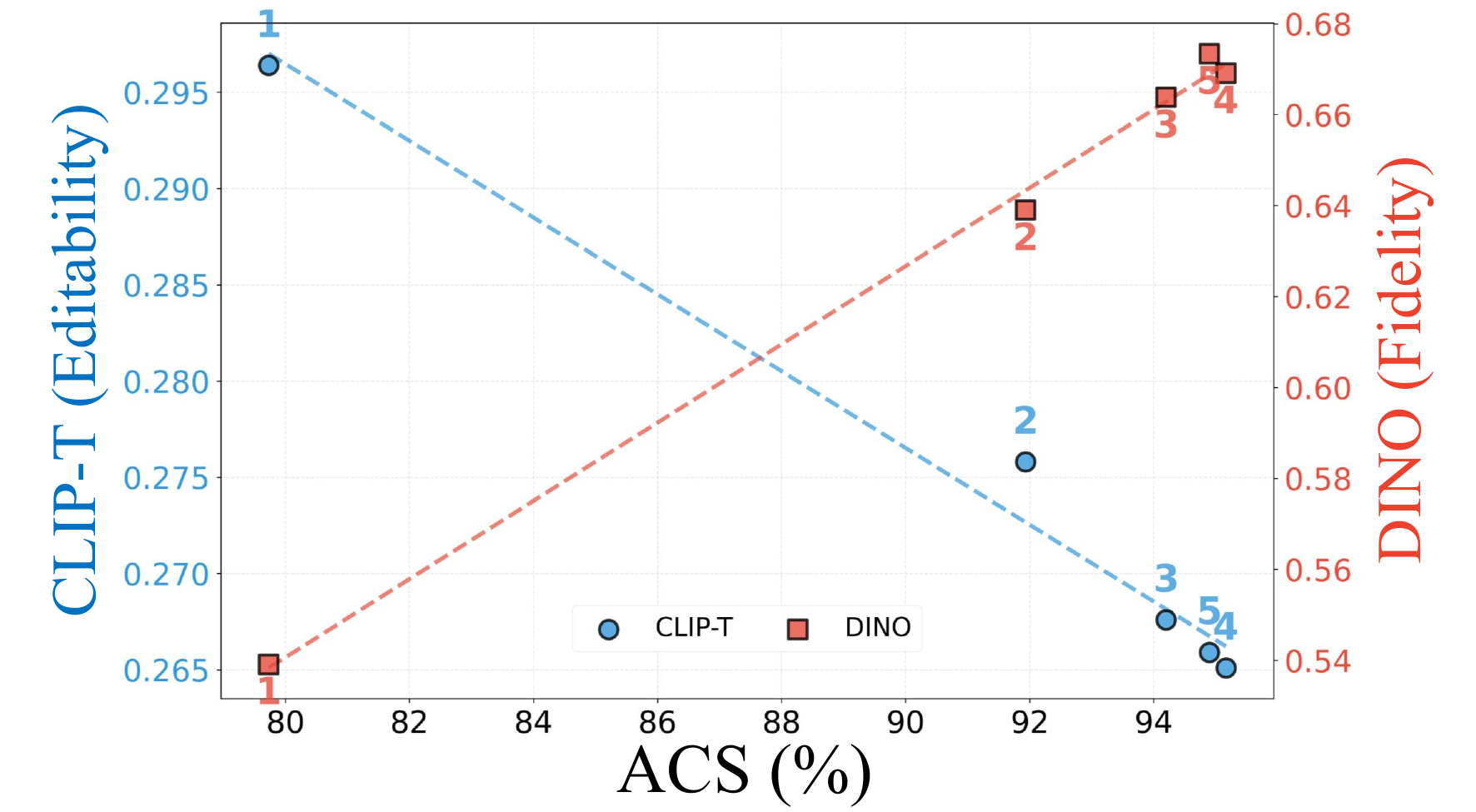}
    \end{subfigure}
    \caption{\textbf{Attention collapse as evidence of conditioning entanglement.}
            Both plots report ACS against fidelity (DINO) and editability (CLIP-T).
            \textbf{(a)} As TI training progresses, ACS rises---fidelity improves but editability degrades accordingly.
            \textbf{(b)} Increasing learnable tokens ($n_{TI}$\,=\,1--5) also raises ACS with the same trade-off, confirming a structural consequence of unified prompts rather than a matter of capacity.}
    \label{fig:collapse_analysis}
\end{figure}

To isolate the causal effect, we treat the pre-trained model as a controlled reference in which the subject position contains a generic class token (e.g., ``dog''), and measure the change when that token is replaced by the learned $S^*$ from TI.
Figure~\ref{fig:attention_analysis} illustrates the analysis results of attention for one prompt example.
With the generic token, attention distributes relatively evenly, allowing the model to integrate subject characteristics with compositional context.
However, substituting a generic token with $S^*$ breaks this balance: the learned token captures most attention (median ACS~$=$~86\%) while context tokens are suppressed to under 10\% combined, with ACS exceeding 80\% in 94\% of cases.

\noindent\textbf{Attention collapse as evidence of conditioning entanglement.}
We next correlate mean ACS with DINO (fidelity) and CLIP-T (editability) for each subject.
As shown in Figure~\ref{fig:collapse_analysis}\textcolor{wacvblue}{(a)}, ACS correlates positively with fidelity and negatively with editability in a consistent, monotonic pattern: as the subject token claims a larger attention share, text-image alignment falls proportionally.
This pattern extends beyond TI: the mean ACS and editability metrics for P+ and NeTI follow the same monotonic ordering, confirming that attention collapse is a directly observable signature of conditioning entanglement.
This analysis demonstrates that attention collapse is a direct and systematic consequence of encoding subject identity through unified prompts in embedding optimization.

\noindent\textbf{Capacity scaling intensifies collapse.}
A natural alternative hypothesis is that the trade-off stems from limited representational capacity: adding more subject tokens $S^*$ could improve fidelity without sacrificing editability~\cite{p+,neti,dciico}.
We test this by training TI, while varying the number of learnable subject tokens ($n_{TI}$) from $1$ to $5$ across our benchmark.
The analysis results falsify the capacity hypothesis.
As shown in Figure~\ref{fig:collapse_analysis}\textcolor{wacvblue}{(b)}, increasing $n_{TI}$ from $1$ to $5$ improves fidelity but degrades editability.
\textcolor{black}{The collective attention share of subject tokens grows markedly, further suppressing context tokens: the added capacity is absorbed by the same shared attention pool, intensifying competition rather than relieving it.}
This intensified dominance, despite being distributed across multiple tokens, \textcolor{black}{indicates that the tension arises from structural competition within the shared attention pool: adding capacity inside a unified prompt intensifies the competition rather than resolving it} (see Section~\ref{sec:supp_capacity_scaling_fig}).
These analysis results rule out capacity as the primary cause of attention competition and reinforce the conclusion from the preceding experiment: subject and context compete within a shared attention space.

\section{Proposed method}
\label{sec:method}

Section~\ref{sec:analysis} establishes that unified prompts force subject identity and scene context to compete for the same attention resources, leading to fidelity--editability trade-off.
More broadly, existing personalization methods across all paradigms route both objectives through a shared conditioning pathway, limiting the ability to independently adjust their balance at inference time.
We present \textit{Decoupled Guidance} (\mymethod{}), a framework that addresses this entanglement at three levels (Figure~\ref{fig:overview}):
(1) contextual co-occurrence in trained tokens (\textit{Context-Agnostic Embeddings}, CAE);
(2) attention competition within a shared forward pass (\textit{Decoupled Guidance Mixer}, DGM);
(3) reference background leakage across spatial regions (\textit{Test-time Cross-Attention Masking}, TCAM).

\begin{figure*}[t]
\centering
\includegraphics[width=1.0\linewidth]{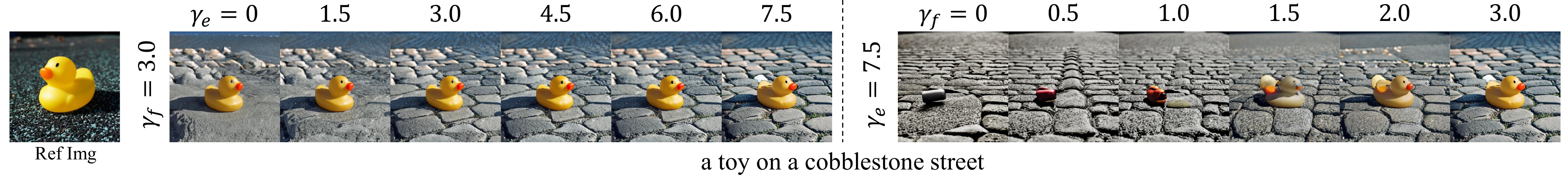}
    \caption{\textbf{Independent guidance control through $\gamma_e$ and $\gamma_f$.}
    Left ($\gamma_f = 3.0$ fixed): Increasing $\gamma_e$ strengthens text adherence while preserving subject identity.
    Right ($\gamma_e = 7.5$ fixed): Increasing $\gamma_f$ enhances subject-specific features while preserving text prompt adherence.
    This validates independent pathway operation.}
\label{fig:supp_guidance_control}
\end{figure*}

\subsection{Preliminaries}
\label{sec:preliminaries}

Our work builds upon Diffusion Models ~\cite{ldm, sdxl, SD3}, which synthesize images by iteratively denoising a latent representation $\mathbf{z}_t$ conditioned on text condition $\vc$ from text encoder $\mathcal{E}$.
Classifier-Free Guidance (CFG)~\cite{cfg} steers generation toward $\vc$ by combining a conditional prediction $\boldsymbol{\epsilon}(\mathbf{z}_t, t, \vc)$ and an unconditional prediction $\boldsymbol{\epsilon}(\mathbf{z}_t)$:
\begin{equation}
\tilde{\boldsymbol{\epsilon}}(\mathbf{z}_t) = \boldsymbol{\epsilon}(\mathbf{z}_t) 
+ \gamma (\boldsymbol{\epsilon}(\mathbf{z}_t, t, \vc) - \boldsymbol{\epsilon}(\mathbf{z}_t)),
\end{equation}
where $t$ denotes the denoising timestep and $\gamma$ controls the guidance strength.
As shown in Section~\ref{sec:analysis}, this unified conditioning forces subject and scene semantics to compete within a single attention space.
We aim to resolve this attention competition by decomposing the conditioning into two parallel streams---subject identity and scene context.

\subsection{Context-Agnostic Embeddings (CAE)}
\label{sec:CAE}
Because standard methods train $S^\ast$ within context-bearing prompts, the resulting embedding already encodes contextual co-occurrence; simply routing it through a separate stream does not eliminate the entanglement.
We therefore introduce \textit{Context-Agnostic Embeddings} (CAE), which trains the token embedding of a new learnable token $[S^\ast]$ in complete isolation.
We use the bracket notation to distinguish our token from $S^\ast$ in existing methods.
Then, the overall prompt sequence consists of only the learnable token $[S^\ast]$ and padding tokens  $[\text{EOT}]$:
\begin{equation}
    [\text{SOT}]\ [S^\ast]\ [\text{EOT}] \cdots [\text{EOT}].
\end{equation}
\textcolor{black}{The token embedding of $[S^\ast]$ is initialized from the embedding of the $[\text{EOT}]$ token and optimized to solely capture subject identity.}
Text encoder $\mathcal{E}$ maps this sequence to the fidelity condition $\vc_f$.
Because no context tokens appear in this sequence, $\vc_f$ is structurally independent of any linguistic context.

\textcolor{black}{Unlike standard methods (e.g., Textual Inversion~\cite{textualinversion}) that train $S^\ast$ within context prompts (e.g., ``\textit{a $S^\ast$ in the snow}''), CAE trains the token embedding of $[S^\ast]$ in isolation with the reconstruction objective:}
\begin{equation}
\mathcal{L} = \mathbb{E}_{\mathbf{z}_t, \boldsymbol{\epsilon}, t} 
\big[\|\boldsymbol{\epsilon} - \boldsymbol{\epsilon}_\theta(\mathbf{z}_t, t, \vc_f)\|^2\big].
\end{equation}

By optimizing this objective,
$\vc_f$ encodes visual patterns from reference images without linguistic grounding, producing a context-independent representation that can compose with arbitrary scene contexts through a separate pathway.
Meanwhile, the context condition $\vc_e$ is obtained by encoding standard text prompts (e.g., ``\textit{a dog in the snow}'') that contain no learnable tokens.

\subsection{Decoupled Guidance Mixer (DGM)}
\label{sec:dgm}
The two conditions $\vc_f$ and $\vc_e$ from CAE each produce a noise prediction, $\boldsymbol{\epsilon}(\mathbf{z}_t, t, \vc_f)$ and $\boldsymbol{\epsilon}(\mathbf{z}_t, t, \vc_e)$, which we incorporate into a factorized reformulation of CFG:
\begin{align}
\label{eq:dgm}
\tilde{\boldsymbol{\epsilon}}(\mathbf{z}_t) &= \boldsymbol{\epsilon}(\mathbf{z}_t) + \gamma_e (\boldsymbol{\epsilon}(\mathbf{z}_t, t, \vc_e) - \boldsymbol{\epsilon}(\mathbf{z}_t)) \nonumber\\
&\quad+ \gamma_f (\boldsymbol{\epsilon}(\mathbf{z}_t, t, \vc_f) - \boldsymbol{\epsilon}(\mathbf{z}_t)),
\end{align}
where $\boldsymbol{\epsilon}(\mathbf{z}_t)$ is the unconditional noise prediction, while $\gamma_e$ and $\gamma_f$ control editability guidance and fidelity guidance, respectively.
Because $\vc_f$ and $\vc_e$ are processed in separate forward passes, each produces its own key--value pairs $\{\mathbf{K}_e, \mathbf{V}_e\}$ and $\{\mathbf{K}_f, \mathbf{V}_f\}$ for the cross-attention mechanism, so the resulting attention maps are disjoint and the pathway-level interference identified in Section~\ref{sec:analysis} is eliminated.
Equation~\ref{eq:dgm} corresponds to the score of a product-of-experts (PoE) target distribution~\cite{Hinton2002PoE,Liu2022ComposableDiffusion} under the conditional independence assumption $\vc_f \!\perp\! \vc_e \mid \mathbf{z}_t$.
This assumption is structurally encouraged by our design: CAE constructs $\vc_f$ without any linguistic tokens and $\vc_e$ contains no learned subject tokens, so the two conditions encode semantically disjoint information.
The full derivation and approximation error analysis are provided in Appendix~\ref{sec:supp_dgm}.


\begin{figure*}[t]
\centering
\includegraphics[width=0.8\textwidth]{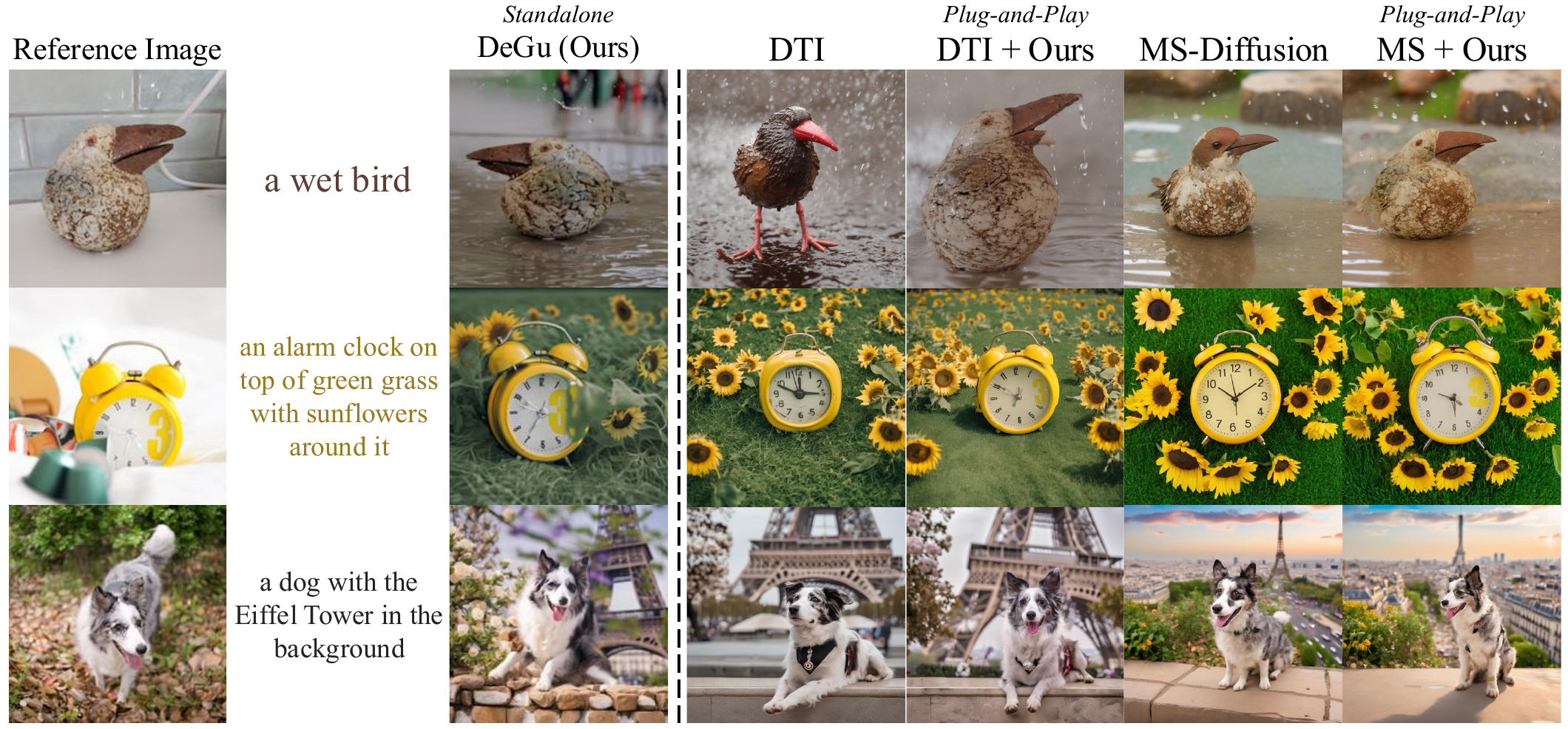}
\caption{
\textbf{Qualitative comparison on SDXL.}
Each row shows one subject: reference (leftmost), standalone \mymethod{}, SDXL baselines (DTI, MS-Diffusion), and plug-and-play variants (DTI+\mymethod{}, MS+\mymethod{}).
DTI follows the prompt but loses subject-specific details; MS-Diffusion retains coarse cues yet misses fine identity details for unseen subjects and fails to compose novel scenes.
\mymethod{} decouples identity and context via dedicated pathways and improves both baselines in a plug-and-play manner without changing their weights.
}
\label{fig:main_qualitative_sdxl}
\end{figure*}

\subsection{Test-time Cross-Attention Masking (TCAM)}
\label{sec:tcam}

DGM eliminates pathway-level interference, but a spatial challenge remains: because $\vc_f$ is trained on reference images, its noise prediction encodes background alongside subject appearance, which can leak into context regions.
We introduce \textit{Test-time Cross-Attention Masking} (TCAM) to spatially partition the two streams at each denoising step, deriving subject masks directly from the latent-space attention within the fidelity forward pass---without any external segmentation model or additional training.
Within the $\vc_f$ forward pass, the CAE tokens $[S^\ast]$ concentrate on subject regions while non-subject tokens ($[\text{SOT}]$, $[\text{EOT}]$) attend to context regions.
Inverting the attention map of such a non-subject token therefore yields more complete and spatially stable subject coverage than relying on $[S^\ast]$ directly.
Let $\mathbf{A}^{\text{ns}}_{t,f}$ denote the cross-attention map of the designated non-subject token at denoising step $t$ in the fidelity stream (see Appendix~\ref{sec:supp_implementation} for implementation details).
We normalize and invert this map, then apply binarization to obtain a subject mask $\mathbf{M}_t$:
\begin{equation}
    \bar{\mathbf{A}}^{\text{ns}}_{t,f} = 1 - \mathrm{Normalize}(\mathbf{A}^{\text{ns}}_{t,f}),
    \label{eq:soft_mask}
\end{equation}
where $\mathrm{Normalize}$ maps values to $[0,1]$.
The soft mask $\bar{\mathbf{A}}^{\text{ns}}_{t,f}$ is binarized via Gaussian blur and Otsu thresholding to yield $\mathbf{M}_t$ (Appendix~\ref{sec:supp_implementation}).
The final guidance is spatially partitioned using $\mathbf{M}_t$:
\begin{align}
\label{eq:tcam_guidance}
\tilde{\boldsymbol{\epsilon}}(\mathbf{z}_t) &= \boldsymbol{\epsilon}(\mathbf{z}_t) \nonumber\\
&\quad+ \gamma_e [(\boldsymbol{\epsilon}(\mathbf{z}_t, t, \vc_e) - \boldsymbol{\epsilon}(\mathbf{z}_t)) \odot (1 - \mathbf{M}_t)] \\
&\quad + \gamma_f [(\boldsymbol{\epsilon}(\mathbf{z}_t, t, \vc_f) - \boldsymbol{\epsilon}(\mathbf{z}_t)) \odot \mathbf{M}_t],\nonumber
\end{align}
confining $\vc_f$ to the subject region and $\vc_e$ to the context region.
TCAM operates entirely within the latent space, reusing attention maps computed during the forward pass, so it adds no external model and no decode overhead (comparison against Grounded SAM 2~\cite{sam2,groundingdino} in Section~\ref{sec:ablation}).

\subsection{Plug-and-play integration}
\label{sec:pnp}
While \mymethod{} can be used with a pretrained backbone model (standalone setting), \mymethod{} can also be seamlessly integrated into existing personalized models with minimal changes.
In practice, an existing personalization method already performs standard CFG with its own condition $\vc_b$ (e.g., ``\textit{a $S^\ast$ in the snow}'').
\mymethod{} appends its fidelity and editability streams alongside $\vc_b$, forming a three-term guidance that combines the existing condition with our decoupled pathways (Appendix~\ref{sec:supp_dgm}, Equations~\ref{eq:supp_dgm_3stream}--\ref{eq:supp_tcam_3stream}).


%

\begin{table*}[t]
\centering
\caption{\textbf{Quantitative comparison across backbones and personalization paradigms.}
\mymethod{} (Ours) consistently pushes the Pareto frontier of personalization across all paradigms and architectures, improving the overall score.
\textbf{CLIP-I} and \textbf{DINO} measure full-image subject fidelity;
\textbf{CLIP-T} measures editability;
\textbf{Overall} $= \frac{1}{4}(\text{CLIP-I} + \text{DINO}) + \frac{1}{2}\text{CLIP-T}$ balances them;
\textbf{MCLIP-I} and \textbf{MDINO} are masked fidelity metrics on the subject region only.}
\label{tab:main_comparison}
\scriptsize
\setlength{\tabcolsep}{3pt}
\renewcommand{\arraystretch}{1.1}
\resizebox{0.8\textwidth}{!}{%
\begin{tabular}{@{}lllcccc@{\hspace{8pt}}cc@{}}
\toprule
\textbf{Backbone} & \textbf{Paradigm} & \textbf{Method}
& CLIP-I $\uparrow$ & DINO $\uparrow$ & CLIP-T $\uparrow$ & \textbf{Overall} $\uparrow$ & MCLIP-I $\uparrow$ & MDINO $\uparrow$ \\

\midrule\midrule 

\multirow{12}{*}{SD v2.1}
 & Standalone & \mymethod{} (Ours)          & 0.7965 & 0.6655 & 0.3034 & \textbf{0.5172} & \textbf{0.9000} & \textbf{0.8019} \\

\cmidrule(l){2-9}
 & \multirow{6}{*}{Embedding}
 & TI~\cite{textualinversion}               & 0.8086 & 0.6256 & 0.2734 & 0.4953 & 0.8763 & 0.7300 \\
 & & \quad + \mymethod{} (Ours)             & 0.8315 & 0.6888 & 0.2870 & \textbf{0.5236} & \textbf{0.9045} & \textbf{0.8086} \\

\cmidrule(l){3-9}
 & & P+~\cite{p+}                           & 0.7828 & 0.6052 & 0.3077 & 0.5009 & 0.8809 & 0.7400 \\
 & & \quad + \mymethod{} (Ours)             & 0.8091 & 0.6719 & 0.3066 & \textbf{0.5236} & \textbf{0.9037} & \textbf{0.8047} \\

\cmidrule(l){3-9}
 & & NeTI~\cite{neti}                       & 0.8092 & 0.6407 & 0.2905 & 0.5077 & 0.8874 & 0.7638 \\
 & & \quad + \mymethod{} (Ours)             & 0.8257 & 0.6885 & 0.2968 & \textbf{0.5270} & \textbf{0.9073} & \textbf{0.8141} \\

\cmidrule(l){2-9}
 & \multirow{4}{*}{Fine-tuning}
 & DB~\cite{dreambooth}                     & 0.8025 & 0.6491 & 0.2996 & 0.5127 & 0.8862 & 0.7792 \\
 & & \quad + \mymethod{} (Ours)             & 0.8272 & 0.6938 & 0.2903 & \textbf{0.5254} & \textbf{0.9066} & \textbf{0.8241} \\

\cmidrule(l){3-9}
 & & CoRe~\cite{core}                       & 0.7428 & 0.5309 & 0.2931 & 0.4650 & 0.8508 & 0.6615 \\
 & & \quad + \mymethod{} (Ours)             & 0.7987 & 0.6514 & 0.3007 & \textbf{0.5129} & \textbf{0.8974} & \textbf{0.7955} \\

\midrule\midrule 

\multirow{6}{*}{SDXL}
 & Standalone & \mymethod{} (Ours)           & 0.7978 & 0.6649 & 0.3021 & \textbf{0.5167} & \textbf{0.9053} & \textbf{0.8065} \\
\cmidrule(l){2-9}
 & Embedding
 & DTI~\cite{dti}                           & 0.7333 & 0.4876 & 0.3105 & 0.4605 & 0.8340 & 0.6212 \\
 & & \quad + \mymethod{} (Ours)             & 0.7800 & 0.6205 & 0.3162 & \textbf{0.5082} & \textbf{0.8941} & \textbf{0.7800} \\

\cmidrule(l){2-9}
 & Encoder
 & MS-Diffusion~\cite{ms-diffusion}         & 0.7655 & 0.5908 & 0.3206 & 0.4994 & 0.8749 & 0.7516 \\
 & & \quad + \mymethod{} (Ours)             & 0.7689 & 0.6072 & 0.3212 & \textbf{0.5046} & \textbf{0.8885} & \textbf{0.7750}\\

\midrule\midrule 

\multirow{3}{*}{SD v3.5}
 & Standalone & \mymethod{} (Ours)           & 0.8029 & 0.6432 & 0.3064 & \textbf{0.5147} & \textbf{0.8817} & \textbf{0.7529} \\
\cmidrule(l){2-9}
 & Fine-tuning
 & DB~\cite{dreambooth}                     & 0.7864 & 0.6250 & 0.3086 & 0.5072 & 0.8744 & 0.7501 \\
 & & \quad + \mymethod{} (Ours)             & 0.8440 & 0.7142 & 0.2852 & \textbf{0.5322} & \textbf{0.9085} & \textbf{0.8214} \\
\bottomrule
\end{tabular}
}
\end{table*}

\section{Experiments}
\label{sec:experiments}
To validate the effectiveness of our proposed framework, \mymethod{}, we address three questions: does pathway separation enable independent control over fidelity and editability across diverse personalization methods and backbones; do the gains stem from the design rather than added capacity; and what is the contribution of each component?


\subsection{Experimental setup}

\noindent\textbf{Datasets and baselines.}
We follow the DreamBooth~\cite{dreambooth} evaluation protocol, using 39 subjects (9 live, 30 non-live) drawn from both the DreamBooth~\cite{dreambooth} and Textual Inversion~\cite{textualinversion} benchmarks to ensure diverse coverage, and evaluate on 25 compositional prompts per subject.
On SD v2.1, we compare against embedding optimization methods (TI~\cite{textualinversion}, P+~\cite{p+}, NeTI~\cite{neti}) and fine-tuning methods (DB~\cite{dreambooth}, CoRe~\cite{core}); on SDXL against DTI~\cite{dti} and MS-Diffusion~\cite{ms-diffusion}; and on SD v3.5 (DiT) we compare against DB-LoRA to test architectural generality.
Full implementation and baseline configuration details are provided in Appendix~\ref{sec:supp_implementation}.

\noindent\textbf{Metrics.}
Following prior work~\cite{dreambooth,textualinversion}, we evaluate subject fidelity using DINO~\cite{dino} and CLIP-I~\cite{clip}, and editability using CLIP-T~\cite{clip}.
We additionally report Masked CLIP-I (MCLIP-I) and Masked DINO (MDINO), which compute fidelity on subject regions cropped via Grounded SAM 2~\cite{sam2,groundingdino} to isolate subject appearance from background changes (see Appendix~\ref{sec:supp_mask_evaluation} for protocol).
We report an overall metric as $\text{Overall} = \frac{1}{4}(\text{CLIP-I} + \text{DINO}) + \frac{1}{2}\text{CLIP-T}$, assigning equal weight to subject fidelity and editability.



\subsection{Experimental results}

\noindent\textbf{Quantitative comparison.}
To evaluate whether pathway separation expands the achievable Pareto frontier regardless of the underlying personalization strategy, Table~\ref{tab:main_comparison} reports results across three backbones (SD v2.1, SDXL, SD v3.5) spanning embedding optimization, fine-tuning, and encoder-based paradigms.
On SD v2.1, \mymethod{} substantially raises subject fidelity across all baselines---MDINO improves by up to $+0.0786$ (TI) and $+0.0449$ (DB)---while maintaining competitive editability, consistently improving the overall score.
These gains extend to SDXL and SD v3.5, which differ substantially in architecture---SD v3.5 replaces the U-Net with a Diffusion Transformer (DiT)---and the fidelity--editability balance remains adjustable at inference time via $\gamma_f$ and $\gamma_e$ (Section~\ref{sec:ablation}).

\begin{figure*}[t]
\centering
\includegraphics[width=0.9\linewidth]{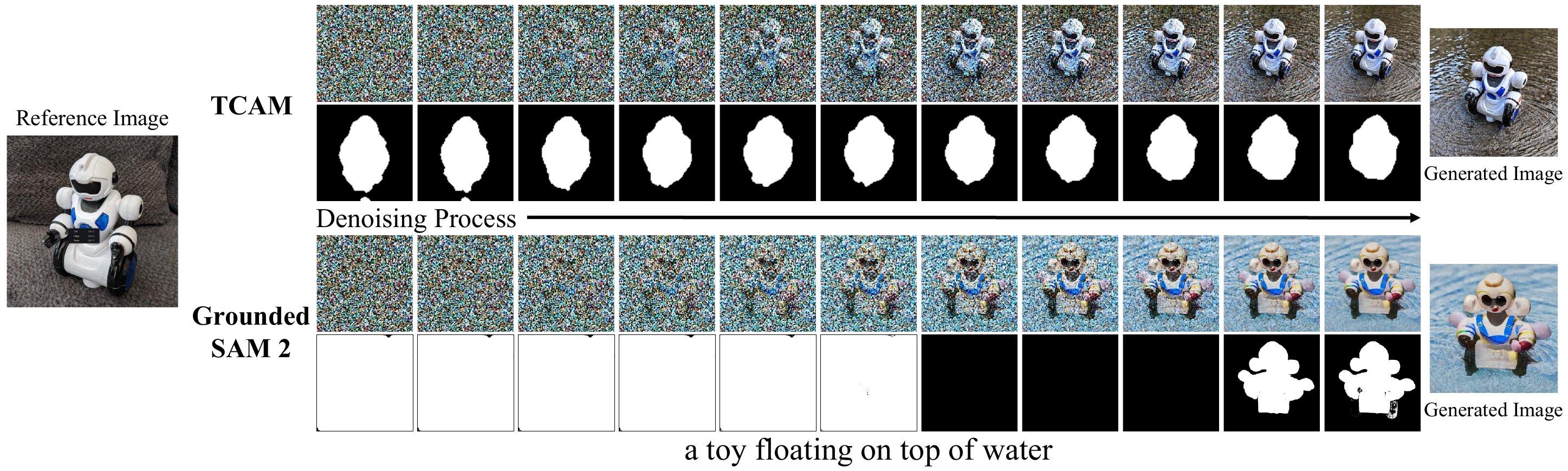}
\caption{\textbf{Mask quality across denoising timesteps.}
Each group shows the decoded intermediate latent and the corresponding mask.
At early timesteps (high noise), Grounded SAM 2 produces unreliable segmentations because its input lies far outside its training distribution, while TCAM consistently localizes the subject by operating on latent-space attention maps.
}
\label{fig:tcam_vs_sam_timestep}
\end{figure*}


\begin{figure}[t]
\centering
\includegraphics[width=1.0\linewidth]{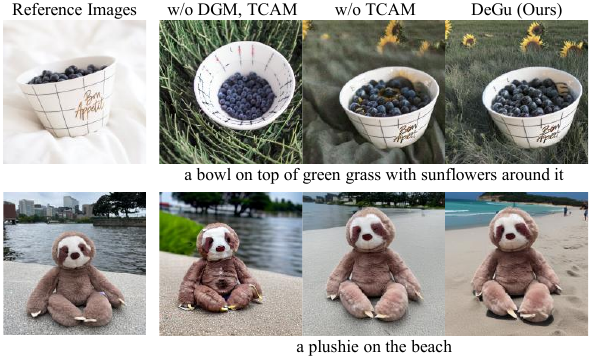}
\caption{\textbf{Qualitative effect of each component.}
\textit{w/o DGM, TCAM}: CAE placed in unified guidance recreates attention collapse.
\textit{w/o TCAM}: DGM decouples guidance but $\vc_f$ spatially dominates, leaking reference backgrounds.
\textit{\mymethod{} (Ours)}: TCAM confines each stream to its relevant region, achieving both high fidelity and compositional control.
}
\label{fig:component_ablation}
\end{figure}

\noindent\textbf{Qualitative comparison.}
Figure~\ref{fig:main_qualitative_sdxl} exposes representative failure modes on SDXL: DTI follows the prompt but loses subject-specific details, while MS-Diffusion retains coarse cues yet misses fine identity for unseen subjects.
\mymethod{} breaks this coupling by routing each objective through a dedicated pathway, and its plug-and-play mode further improves each baseline without backbone-specific retraining.
The same pattern holds on SD v2.1, where embedding optimization methods lose subject identity and fine-tuning methods fail to compose novel scenes (Appendix~\ref{sec:supp_sd21_qualitative}).
SD v3.5 results are shown in Figure~\ref{fig:results_summary}, with additional examples in Appendix~\ref{sec:supp_sd35_results}.


\noindent\textbf{Plug-and-play integration.}
Improving an existing method without degrading compositional capability is structurally difficult under entangled conditioning, because any added fidelity signal injected into a shared attention stream competes with the host context tokens.
Table~\ref{tab:main_comparison} (rows marked $+$\,\mymethod{}) shows that subject fidelity improves substantially across tested baselines while the overall score improves consistently, with only a few hundred kilobytes of storage for the CAE token embeddings (details in Appendix~\ref{sec:supp_size}).
Crucially, CAE token embeddings are learned on the pretrained backbone and then reused in all comparison settings, rather than being optimized separately for each baseline.


\subsection{Ablation studies}
\label{sec:ablation}

Having established that each component targets a distinct level of entanglement, including conditioning-level competition (CAE), pathway-level interference (DGM), and spatial-level leakage (TCAM), we now verify their individual contributions through systematic ablations on SD v2.1.

\begin{table}[t!]
\centering
\caption{\textbf{Component ablation.}
Each row adds one component.
We report masked fidelity metrics (MCLIP-I, MDINO) to isolate subject appearance from background changes (Overall $= \frac{1}{4}(\text{MCLIP-I} + \text{MDINO}) + \frac{1}{2}\text{CLIP-T}$).
CAE alone recreates attention collapse; DGM recovers fidelity but leaks reference backgrounds; TCAM resolves this by spatially confining each stream.
}
\label{tab:component_ablation}
\small
\resizebox{1.0\linewidth}{!}{
\setlength{\tabcolsep}{4pt}
\begin{tabular}{@{}ccc ccc c@{}}
\toprule
CAE & DGM & TCAM & MCLIP-I$\uparrow$ & MDINO$\uparrow$ & CLIP-T$\uparrow$ & Overall$\uparrow$ \\
\midrule
\cmark & \xmark & \xmark & 0.8897 & 0.7683 & 0.2751 & 0.5521 \\
\cmark & \cmark & \xmark & 0.8956 & 0.7902 & 0.2995 & 0.5712 \\
\cmark & \cmark & \cmark & \textbf{0.9000} & \textbf{0.8019} & \textbf{0.3034} & \textbf{0.5772} \\
\bottomrule
\end{tabular}
}
\end{table}

\noindent\textbf{Component analysis.}
Table~\ref{tab:component_ablation} and Figure~\ref{fig:component_ablation} report results for three progressive configurations.
CAE alone recreates attention collapse when placed back into unified guidance, degrading both metrics.
Adding DGM decouples the guidance streams but causes $\vc_f$ to dominate spatially, leaking reference backgrounds into generated scenes.
TCAM resolves this by spatially partitioning the two streams, recovering editability while maintaining high fidelity.

\noindent\textbf{Why not external segmentation?}
External segmenters such as Grounded SAM 2~\cite{sam2,groundingdino} require decoding the noisy latent $\mathbf{z}_t$ to pixel space at every step; at high-noise timesteps, the decoded output is heavily out-of-distribution, producing unstable masks.
As shown in Figure~\ref{fig:tcam_vs_sam_timestep}, external segmentation fails under high noise, while cross-attention maps still localize the subject.
TCAM achieves higher subject fidelity (MDINO: 0.8019 vs.\ 0.6214) and faster inference (4.5\,s/image vs.\ 21.2\,s/image).

\noindent\textbf{Inference-time guidance control.}
Because $\vc_f$ and $\vc_e$ operate through separate forward passes with distinct cross-attention mechanisms, the fidelity--editability balance becomes adjustable at inference time without retraining.
Figure~\ref{fig:supp_guidance_control} illustrates this: increasing $\gamma_e$ (with $\gamma_f = 3.0$ fixed) strengthens text adherence while preserving subject identity, and increasing $\gamma_f$ (with $\gamma_e = 7.5$ fixed) sharpens subject-specific features while compositional elements remain stable.
Entangled methods lack this capability.


\section{Conclusion}
We identify \textit{conditioning entanglement}---routing subject identity and scene context through a shared pathway---as a structural bottleneck that collapses compositional controllability in text-to-image personalization, with \textit{attention collapse} as its observable signature.
\textit{Decoupled Guidance} (\mymethod{}) addresses this at three levels: Context-Agnostic Embeddings (CAE) train the subject token in a padding-only sequence to eliminate linguistic co-occurrence; the Decoupled Guidance Mixer (DGM) separates subject and context conditioning with disjoint key--value sets; and Test-time Cross-Attention Masking (TCAM) spatially confines each stream via a binary mask derived from the fidelity-stream attention maps, without requiring an external segmentation module.
By decomposing guidance into independent streams, \mymethod{} shifts the fidelity--editability balance from a training-time decision to an inference-time choice, enabling flexible control via $\gamma_f$ and $\gamma_e$.
Across embedding optimization, fine-tuning, and encoder-based methods on multiple backbones, this design consistently improves the overall personalization performance, demonstrating the flexibility and effectiveness of the proposed decoupled guidance scheme.

\clearpage
\appendix
\phantomsection
\pdfbookmark[1]{Supplementary Material}{supplementary}
\section*{Supplementary Material}
\setcounter{page}{1}
\setcounter{figure}{0}
\setcounter{table}{0}
\setcounter{equation}{0}
\setcounter{section}{0}
\renewcommand{\thesection}{\Alph{section}}
\renewcommand{\theHsection}{supp.\Alph{section}}
\renewcommand{\thefigure}{\Alph{figure}}
\renewcommand{\theHfigure}{supp.\Alph{figure}}
\renewcommand{\thetable}{\Alph{table}}
\renewcommand{\theHtable}{supp.\Alph{table}}
\renewcommand{\theequation}{\Alph{equation}}
\renewcommand{\theHequation}{supp.\Alph{equation}}

\section*{Table of Contents}
\newcommand{\tocmain}[3]{\noindent\makebox[1.2em][r]{\textbf{\hyperref[#1]{#2}}}\hspace{6pt}#3\dotfill\pageref{#1}\par\medskip}
\newcommand{\tocsub}[3]{\noindent\hspace{1.2em}\hspace{6pt}\hyperref[#1]{#2}~~#3\dotfill\pageref{#1}\par}
\newcommand{\tocsubend}[3]{\noindent\hspace{1.2em}\hspace{6pt}\hyperref[#1]{#2}~~#3\dotfill\pageref{#1}\par\medskip}

\tocmain{sec:supp_dgm}{A}{Theoretical justification for the Decoupled Guidance Mixer}
\tocmain{sec:supp_implementation}{B}{Implementation details}
\tocsub{sec:supp_train_config}{B.1}{Training configurations}
\tocsub{sec:supp_baseline_config}{B.2}{Baseline configurations}
\tocsubend{sec:supp_eval_config}{B.3}{Evaluation configuration}
\tocmain{sec:supp_mask_evaluation}{C}{Mask-based evaluation protocols}
\tocmain{sec:supp_extended_quant}{D}{Additional quantitative results}
\tocmain{sec:supp_additional_results}{E}{Additional qualitative results}
\tocsub{sec:supp_sd21_qualitative}{E.1}{Qualitative comparison on SD v2.1}
\tocsub{sec:supp_sd35_results}{E.2}{Results with SD v3.5 backbone}
\tocsubend{sec:supp_plugandplay}{E.3}{Plug-and-play integration on SD v2.1}
\tocmain{sec:supp_user_study}{F}{User study}
\tocmain{sec:supp_supple_ablation}{G}{Additional ablation studies}
\tocsub{sec:supp_capacity_scaling_fig}{G.1}{Token capacity scaling}
\tocsub{sec:supp_ablation_cfe_tokens}{G.2}{CAE token count}
\tocsub{sec:supp_ablation_masking}{G.3}{Masking token selection}
\tocsub{sec:supp_disentangle}{G.4}{Disentanglement as the source of gain in plug-and-play mode}
\tocsubend{sec:supp_size}{G.5}{Storage-performance trade-off}
\tocmain{sec:supp_multi_concept}{H}{Multi-concept personalization}

\noindent
This supplementary material provides theoretical derivations, implementation details, extended quantitative and qualitative results, user study protocols, ablation studies, and multi-concept experiments that complement the main paper.


\section{Theoretical justification for the Decoupled Guidance Mixer}
\label{sec:supp_dgm}

The Decoupled Guidance Mixer (DGM, Section~\ref{sec:dgm}) combines two independent noise predictions via
\begin{equation}
\begin{split}
\tilde{\boldsymbol{\epsilon}}(\mathbf{z}_t)
&= \boldsymbol{\epsilon}(\mathbf{z}_t, t) \\
&\quad + \gamma_e\big(\boldsymbol{\epsilon}(\mathbf{z}_t,t,\vc_e)
       - \boldsymbol{\epsilon}(\mathbf{z}_t, t)\big) \\
&\quad + \gamma_f\big(\boldsymbol{\epsilon}(\mathbf{z}_t,t,\vc_f)
       - \boldsymbol{\epsilon}(\mathbf{z}_t, t)\big).
\end{split}
\label{eq:supp_dgm}
\end{equation}
A natural question is whether this linear combination has a principled probabilistic interpretation, or whether it is a heuristic superposition of two guidance signals.
A product-of-experts (PoE) distribution~\cite{Hinton2002PoE} assigns high density only where every expert agrees, which aligns with the goal of personalization: the generated image should satisfy both the fidelity condition and the editability condition simultaneously.
We show that Equation~\ref{eq:supp_dgm} corresponds to the score of such a PoE target, and that it approximates the joint posterior score with an explicit error term.

\paragraph{Score function notation.}
In the main paper, the unconditional prediction is written as $\boldsymbol{\epsilon}(\mathbf{z}_t)$ to visually distinguish it from the conditional prediction $\boldsymbol{\epsilon}(\mathbf{z}_t, t, \vc)$, since $t$ is already implied by the subscript of $\mathbf{z}_t$.
Throughout this section, we write $\boldsymbol{\epsilon}(\mathbf{z}_t, t)$ for both cases to keep the timestep dependence explicit in the derivations.
Define the (conditional) score function
\begin{equation}
s(\mathbf{z}_t, t\mid \vc)\triangleq \nabla_{\mathbf{z}_t}\log p_t(\mathbf{z}_t\mid \vc).
\end{equation}
Under the standard $\boldsymbol{\epsilon}$-prediction parameterization~\cite{ddpm}, the score is proportional to the negative predicted noise:
\begin{equation}
s_{\theta}(\mathbf{z}_t,t\mid \vc)\;=\;-\frac{1}{\sigma_t}\boldsymbol{\epsilon}_{\theta}(\mathbf{z}_t,t,\vc),
\end{equation}
where $\sigma_t$ is the noise scale at time $t$.

\paragraph{A target distribution whose score factorizes additively.}
Rather than claiming that a linear combination of scores always equals the true joint score of $p_t(\mathbf{z}_t\mid \vc_f,\vc_e)$, we define an explicit target distribution whose score decomposes additively.
Consider the following product-of-experts distribution~\cite{Hinton2002PoE,Liu2022ComposableDiffusion} with $\gamma_f=\gamma_e=1$:
\begin{equation}
\label{eq:supp_poe}
q(\mathbf{z}_t\mid \vc_f,\vc_e)
\;\propto\;
p_t(\mathbf{z}_t)\;
p(\vc_f\mid \mathbf{z}_t)\;
p(\vc_e\mid \mathbf{z}_t).
\end{equation}
Taking the gradient of the log-density:
\begin{equation}
\begin{split}
\nabla_{\mathbf{z}_t}\log q(\mathbf{z}_t\mid \vc_f,\vc_e)
&= \nabla_{\mathbf{z}_t}\log p_t(\mathbf{z}_t) \\
&\quad + \nabla_{\mathbf{z}_t}\log p(\vc_f\mid \mathbf{z}_t) \\
&\quad + \nabla_{\mathbf{z}_t}\log p(\vc_e\mid \mathbf{z}_t).
\end{split}
\label{eq:supp_poe_score_step1}
\end{equation}
Applying Bayes' rule and noting that $\nabla_{\mathbf{z}_t}\log p(\vc)=0$:
\begin{equation}
\begin{split}
\nabla_{\mathbf{z}_t}\log p(\vc\mid \mathbf{z}_t)
&= \nabla_{\mathbf{z}_t}\log p_t(\mathbf{z}_t\mid \vc) \\
&\quad - \nabla_{\mathbf{z}_t}\log p_t(\mathbf{z}_t),
\end{split}
\end{equation}
we obtain
\begin{equation}
\begin{split}
\nabla_{\mathbf{z}_t}\log q(\mathbf{z}_t\mid \vc_f,\vc_e)
&= s(\mathbf{z}_t, t) \\
&\quad + \big(s(\mathbf{z}_t, t\mid \vc_f)
       - s(\mathbf{z}_t, t)\big) \\
&\quad + \big(s(\mathbf{z}_t, t\mid \vc_e)
       - s(\mathbf{z}_t, t)\big),
\end{split}
\label{eq:supp_poe_score_step2}
\end{equation}
where $s(\mathbf{z}_t, t)\triangleq \nabla_{\mathbf{z}_t}\log p_t(\mathbf{z}_t)$ is the unconditional score.
Substituting $s_{\theta}(\mathbf{z}_t, t\mid c)=-\boldsymbol{\epsilon}_{\theta}(\mathbf{z}_t, t, c)/\sigma_t$ shows that this special case ($\gamma_f=\gamma_e=1$) recovers the DGM update in Equation~\ref{eq:supp_dgm} with unit guidance scales.
DGM generalizes this by introducing independent guidance scales $\gamma_f$ and $\gamma_e$, corresponding to an exponentiated PoE target $q_\gamma(\mathbf{z}_t \mid \vc_f, \vc_e) \propto p_t(\mathbf{z}_t)\, p(\vc_f \mid \mathbf{z}_t)^{\gamma_f}\, p(\vc_e \mid \mathbf{z}_t)^{\gamma_e}$, following the same convention as standard CFG~\cite{cfg} where $\gamma>1$ sharpens the conditional distribution.
The effect of each scale on the two streams is validated independently in Figure~\ref{fig:supp_guidance_control} of the main paper.

\paragraph{Approximation error and its suppression.}
When the two conditions are not conditionally independent given $\mathbf{z}_t$, define the interaction residual
\begin{equation}
\label{eq:supp_interaction}
\begin{split}
\Delta(\mathbf{z}_t)
&\triangleq \log p(\vc_f,\vc_e\mid \mathbf{z}_t) \\
&\quad -\log p(\vc_f\mid \mathbf{z}_t)
-\log p(\vc_e\mid \mathbf{z}_t),
\end{split}
\end{equation}
so that the true joint score decomposes as
\begin{equation}
\begin{split}
\nabla_{\mathbf{z}_t}\log p_t(\mathbf{z}_t\mid \vc_f,\vc_e)
&= \nabla_{\mathbf{z}_t}\log q(\mathbf{z}_t\mid \vc_f,\vc_e) \\
&\quad +\nabla_{\mathbf{z}_t}\Delta(\mathbf{z}_t).
\end{split}
\end{equation}
DGM can therefore be viewed as an approximation to the true joint score, with $\nabla_{\mathbf{z}_t}\Delta(\mathbf{z}_t)$ as an explicit error term.
It is worth distinguishing the consensus mechanism inherent to PoE from the problematic competition observed in conventional single-pass conditioning.
In a PoE distribution, the requirement that every expert assigns high likelihood is a desirable property: it ensures that the generated sample simultaneously satisfies both the fidelity and editability objectives.
The difficulty arises when this consensus is realized \emph{implicitly} inside a single forward pass, where the conditions share a common softmax and the resulting attention competition corrupts each expert's likelihood estimate---the attention collapse documented in Section~\ref{sec:analysis}.
DGM relocates the consensus from the attention interior to the score space: each expert produces its likelihood estimate in an isolated forward pass, and the estimates are combined externally via Equation~\ref{eq:supp_dgm}.
Our design further suppresses the approximation error $\nabla_{\mathbf{z}_t}\Delta(\mathbf{z}_t)$ in two complementary ways:
(i) CAE removes shared linguistic context tokens from $\vc_f$ by construction, reducing statistical dependence between $\vc_f$ and $\vc_e$; and
(ii) because the two forward passes maintain disjoint key-value pools, each expert estimates its conditional likelihood without computational interference from the other stream.
Together, (i) and (ii) make the PoE approximation most accurate in the regime where $\Delta(\mathbf{z}_t)$ is small, i.e., when $\vc_f$ and $\vc_e$ are close to conditionally independent given $\mathbf{z}_t$.

\paragraph{Scope of the guarantee: single-step isolation.}
The pathway isolation established above holds \emph{within a single denoising step}: given $\mathbf{z}_t$, the two forward passes are independent.
However, across steps, $\mathbf{z}_{t-1}$ is produced by combining both guidance signals (Equation~\ref{eq:supp_dgm}), so $\mathbf{z}_{t-1}$ carries information from both $\vc_f$ and $\vc_e$.
At the next step, the fidelity forward pass $\boldsymbol{\epsilon}(\mathbf{z}_{t-1},t{-}1,\vc_f)$ therefore receives an indirect signal from $\vc_e$ through $\mathbf{z}_{t-1}$.
This inter-step coupling is a structural property of all score-level composition methods: Composable Diffusion~\cite{Liu2022ComposableDiffusion}, which combines multiple text conditions via the same additive formulation, shares exactly the same inter-step dependency but does not discuss it explicitly.
Moreover, this inter-step mixing is not merely a limitation---it serves a constructive role.
If the two streams were fully decoupled across steps, each would independently develop its own spatial structure, causing the subject to appear as a visually inconsistent element pasted onto the scene rather than naturally integrated into it.
By sharing $\mathbf{z}_{t-1}$ across streams, DGM ensures that the fidelity and editability streams develop a common spatial scaffold, allowing the subject and scene to remain geometrically coherent throughout denoising.
The key advantage of DGM is therefore that it removes \emph{intra-step} coupling---where $\vc_f$ and $\vc_e$ would otherwise compete within the same forward pass under a shared softmax---while retaining inter-step mixing as a mechanism for spatial coherence.
Our ablation results (Table~\ref{tab:component_ablation} in the main paper) confirm that intra-step isolation is the dominant factor driving the improvement.

\paragraph{Extension to three-stream plug-and-play integration.}
When \mymethod{} is attached to an existing personalization method (Section~\ref{sec:pnp}), a third condition $\vc_b$ (the prompt containing the learned token of the host method, e.g., ``\textit{a $S^*$ in the snow}'') enters the guidance.
The three-stream DGM update without spatial masking takes the form
\begin{align}
\label{eq:supp_dgm_3stream}
\tilde{\boldsymbol{\epsilon}}(\mathbf{z}_t) 
&= \boldsymbol{\epsilon}(\mathbf{z}_t, t) \nonumber\\
&\quad + \gamma_e\big(\boldsymbol{\epsilon}(\mathbf{z}_t,t,\vc_e)-\boldsymbol{\epsilon}(\mathbf{z}_t, t)\big) \nonumber \\
&\quad + \gamma_f\big(\boldsymbol{\epsilon}(\mathbf{z}_t,t,\vc_f)-\boldsymbol{\epsilon}(\mathbf{z}_t, t)\big) \\
&\quad + \gamma_b\big(\boldsymbol{\epsilon}(\mathbf{z}_t,t,\vc_b)-\boldsymbol{\epsilon}(\mathbf{z}_t, t)\big). \nonumber
\end{align}
When combined with TCAM, the spatial partitioning confines $\vc_f$ to the subject region and $\vc_e$ to the background, while $\vc_b$ operates over the full spatial extent since it already encodes both subject and context in its entangled prompt:
\begin{align}
\label{eq:supp_tcam_3stream}
\tilde{\boldsymbol{\epsilon}}(\mathbf{z}_t)
&= \boldsymbol{\epsilon}(\mathbf{z}_t, t) \nonumber \\
&\quad + \gamma_e \big[(\boldsymbol{\epsilon}(\mathbf{z}_t,t,\vc_e) - \boldsymbol{\epsilon}(\mathbf{z}_t, t)) \odot (1 - \mathbf{M}_t)\big] \nonumber \\
&\quad + \gamma_f \big[(\boldsymbol{\epsilon}(\mathbf{z}_t,t,\vc_f) - \boldsymbol{\epsilon}(\mathbf{z}_t, t)) \odot \mathbf{M}_t\big] \\
&\quad + \gamma_b \big(\boldsymbol{\epsilon}(\mathbf{z}_t,t,\vc_b) - \boldsymbol{\epsilon}(\mathbf{z}_t, t)\big). \nonumber
\end{align}
The PoE framework extends naturally to three experts.
Define the three-expert target distribution:
\begin{equation}
\label{eq:supp_poe3}
q(\mathbf{z}_t \mid \vc_f, \vc_e, \vc_b)
\;\propto\;
p_t(\mathbf{z}_t)\;
p(\vc_f \mid \mathbf{z}_t)\;
p(\vc_e \mid \mathbf{z}_t)\;
p(\vc_b \mid \mathbf{z}_t).
\end{equation}
Taking the log-gradient and applying Bayes' rule yields
\begin{align}
\label{eq:supp_poe3_score}
\nabla_{\mathbf{z}_t} \log q
&= s(\mathbf{z}_t, t) \notag \\
&\quad + \big(s(\mathbf{z}_t, t \mid \vc_f) - s(\mathbf{z}_t, t)\big) \notag \\
&\quad + \big(s(\mathbf{z}_t, t \mid \vc_e) - s(\mathbf{z}_t, t)\big) \notag \\
&\quad + \big(s(\mathbf{z}_t, t \mid \vc_b) - s(\mathbf{z}_t, t)\big).
\end{align}
which, under the $\boldsymbol{\epsilon}$-parameterization, corresponds to Equation~\ref{eq:supp_dgm_3stream} with unit guidance scales.
The interaction residual generalizes to
\begin{equation}
\begin{split}
\Delta(\mathbf{z}_t)
&= \log p(\vc_f, \vc_e, \vc_b \mid \mathbf{z}_t)
- \log p(\vc_f \mid \mathbf{z}_t) \\
& \quad - \log p(\vc_e \mid \mathbf{z}_t) 
- \log p(\vc_b \mid \mathbf{z}_t).
\end{split}
\label{eq:supp_interaction3}
\end{equation}
The same suppression mechanisms from the two-expert case---CAE and disjoint forward passes---apply here.
Setting $\gamma_b = 0$ recovers the standalone two-expert form (Equation~\ref{eq:supp_dgm}), confirming that the three-expert formulation is a strict generalization.


\section{Implementation details}
\label{sec:supp_implementation}

Throughout the supplementary material, we follow the notation convention from the main paper: $[S^*]$ denotes the learnable token introduced by our method (Context-Agnostic Embeddings), while $S^*$ refers to learned tokens in existing methods such as Textual Inversion.

\subsection{Training configurations}
\label{sec:supp_train_config}
\noindent\textbf{Denoising module architecture.}
Throughout this paper, we refer to the core network that predicts noise at each denoising step as the \textit{denoising module}.
For SD v1.5, SD v2.1, and SDXL, this module is a U-Net with explicit cross-attention layers.
For SD v3.5, the denoising module is a Diffusion Transformer (DiT) that employs joint attention (MMDiT), where text and image tokens are concatenated and processed through shared self-attention blocks rather than separate cross-attention layers.
In the latter case, TCAM extracts the text-token-to-image-patch submatrix from the joint attention as the functional equivalent of a cross-attention map.
All backbone-specific details---including attention source, non-subject token selection, candidate resolution sets for mask generation, and guidance scales---are reported in the following subsections.

\noindent\textbf{Plug-and-play design.}
Our method is designed as a modular framework that can be integrated with existing personalization methods without modifying their base architectures.
The core components (CAE, DGM, TCAM) operate independently of the underlying diffusion backbone and can be applied to embedding optimization, fine-tuning, and encoder-based approaches alike.

\noindent\textbf{Optimization setup.}
We train CAE tokens using AdamW optimizer with learning rate $\eta = 1\text{e-}3$ and weight decay $1\text{e-}2$.
Training is performed for 1500 steps with batch size 8.
We use a constant learning rate schedule and further scale the learning rate by the batch size.
\textcolor{black}{For experiments with SD v3.5, we instead use the Prodigy optimizer.}

\noindent\textbf{CAE.}
Unless specified otherwise, we use $5$ context-agnostic embeddings with layer-wise conditioning throughout the cross-attention layers following P+, providing favorable fidelity while maintaining storage efficiency (320\,KB vs.\ 3.4\,GB for fine-tuning methods, e.g., for SD v2.1), as discussed in Section~\ref{sec:supp_ablation_cfe_tokens} and~\ref{sec:supp_size}.
When multiple CAE tokens exist, $\mathbf{A}^{\text{ns}}_{t,f}$ refers to the attention map of the single non-subject token (e.g., $[\text{SOT}]$, $[\text{EOT}]$), while the subject token attention is computed as the average: $\mathbf{A}^{[S^*]}_{t,f} = \frac{1}{n}\sum_{i=1}^{n} \mathbf{A}^{[S^*]}_{t,f,i}$, where $i$ indexes the CAE tokens and $n$ is their count.

\noindent\textbf{TCAM activation timing.}
We activate TCAM when the denoising timestep satisfies $t < t_{\text{TCAM}}$, where $t_{\text{TCAM}} = 0.6T$ and $T$ is the total number of denoising steps.
During the early phase of denoising ($t \geq t_{\text{TCAM}}$), the two guidance streams operate without spatial partitioning so that they jointly establish coherent global structure; once coarse layout is formed ($t < t_{\text{TCAM}}$), TCAM activates to confine each stream to its semantically relevant region, refining subject details without leaking reference backgrounds.
We select $t_{\text{TCAM}} = 0.6T$ based on empirical evaluation across subjects, where fidelity gains saturate while editability remains stable.

\textcolor{black}{\noindent\textbf{TCAM configuration per backbone.}
TCAM requires two backbone-dependent choices: (i) the attention source and the non-subject token used for mask derivation, and (ii) the candidate set used for mask generation.
In SD v1.5 and SD v2.1, which employ a U-Net with explicit cross-attention layers, we extract cross-attention maps directly from the decoder blocks.
For mask derivation, we use the $[\text{SOT}]$ (start-of-text) token as the non-subject token, which serves as a sequence-level aggregator and consistently attends to background regions (see ablation in Section~\ref{sec:supp_ablation_masking}).
In SDXL, we likewise extract cross-attention maps from the decoder blocks of the U-Net architecture.
However, the non-subject token used for mask derivation is $[\text{EOT}]$ (end-of-text), whose attention distribution more consistently reflects global context and background regions.
In SD v3.5, which employs a Diffusion Transformer (DiT) with joint attention (MMDiT), no separate cross-attention layer exists; instead, text and image tokens are concatenated and processed through shared self-attention blocks.
We extract the submatrix corresponding to the text-token-to-image-patch direction from the joint attention matrix and treat it as the functional equivalent of a cross-attention map.
The non-subject token in this architecture is the sequence-terminal token $[\text{EOT}]$, which plays a similar global aggregation role in the text representation.
The soft mask is computed by inverting the selected non-subject token attention: $\bar{\mathbf{A}}^{\text{ns}}_{t,f} = 1 - \mathrm{Normalize}(\mathbf{A}^{\text{ns}}_{t,f})$, where $\mathbf{A}^{\text{ns}}_{t,f}$ denotes the attention map of the non-subject token in the fidelity stream at denoising step $t$.
We then apply Gaussian blur with kernel size $5\times5$ and sigma $1.1$, followed by Otsu thresholding to binarize $\bar{\mathbf{A}}^{\text{ns}}_{t,f}$ into the final mask $\mathbf{M}_t$.
The candidate resolution or layer set is chosen per backbone.
For SD v1.5, we select from resolutions $\{16, 32\}$, and for SD v2.1 from $\{24, 48\}$; in each case, the candidate mask with lower Shannon entropy is selected.
For SDXL, we use a fixed resolution of $32$.
For SD v3.5, we average the attention maps from the middle transformer layers $\{8, 9, 10\}$.
}

\noindent\textbf{Guidance scales.}
We use $\gamma_f = 3.0$ for the fidelity stream and $\gamma_e = 7.5$ for the editability stream in standalone mode.
We set $\gamma_e = 7.5$ to match the default CFG scale of pre-trained Stable Diffusion, thereby preserving the original compositional capabilities.
For the fidelity stream, we empirically select $\gamma_f = 3.0$ as a moderate value that enhances subject identity without overwhelming the editability guidance.
In plug-and-play mode (Equation~\ref{eq:supp_tcam_3stream}), we use $\gamma_b = 4.0$, $\gamma_f = 1.5$, and $\gamma_e = 3.0$.

\subsection{Baseline configurations}
\label{sec:supp_baseline_config}
We compare against representative methods from three paradigms: embedding optimization (Textual Inversion~\cite{textualinversion}, P+~\cite{p+}, NeTI~\cite{neti}, DTI~\cite{dti}), fine-tuning (DreamBooth~\cite{dreambooth}, CoRe~\cite{core}), and encoder-based methods (MS-Diffusion~\cite{ms-diffusion}).
All baselines follow their official implementations when available; for SD v3.5, we adopt a DreamBooth-style LoRA adaptation of the denoising backbone.

\noindent\textit{SD v1.5 / SD v2.1 baselines:}
\begin{itemize}
    \item \textbf{Textual Inversion:} Single learnable token, 3000 training steps, learning rate 5e-3, batch size 8.
    \item \textbf{P+:} Extended token sequence ($n=16$), 500 steps, learning rate 1e-3, batch size 8.
    \item \textbf{NeTI:} Timestep-dependent embeddings, 1000 steps, learning rate 1e-3, batch size 8.
    \item \textbf{DreamBooth:} Full U-Net fine-tuning, 800 steps, learning rate 5e-4, batch size 1.
    \item \textbf{CoRe:} Context-regularized training for embedding (300 steps, learning rate 5e-3, batch size 6) and full U-Net fine-tuning (1000 steps, learning rate 2e-6, batch size 4).
\end{itemize}

\noindent\textit{SDXL baselines:}
\begin{itemize}
    \item \textbf{DTI:} Single learnable token, 800 training steps, learning rate 1e-2, batch size 4.
    \item \textbf{MS-Diffusion:} Zero-shot inference using officially released pre-trained weights; no per-subject optimization is required by design.
\end{itemize}

\noindent\textit{SD v3.5 baselines:}
For brevity, this SD v3.5 DreamBooth-style LoRA baseline is denoted as DB in Table~\ref{tab:main_comparison} of the main paper.
\begin{itemize}
    \item \textbf{DB-LoRA:} LoRA fine-tuning of the SD v3.5 denoising backbone (MMDiT, rank 8), 800 steps, Prodigy optimizer, batch size 4.
\end{itemize}

\subsection{Evaluation configuration}
\label{sec:supp_eval_config}
\noindent\textbf{Datasets.}
We follow the DreamBooth evaluation protocol~\cite{dreambooth}, using 39 subjects: 9 live subjects (e.g., dog, cat) and 30 non-live objects (e.g., backpack, teapot, clock, toy), drawn from both the DreamBooth~\cite{dreambooth} and Textual Inversion~\cite{textualinversion} benchmarks.
For each subject, we use 4--6 reference images and evaluate on 25 compositional prompts from the DreamBooth benchmark.

\noindent\textbf{Prompt templates.}
We evaluate our method using all 25 prompts from the DreamBooth benchmark~\cite{dreambooth}.
The benchmark provides distinct prompt sets for non-live subjects and live subjects.
Non-live prompts emphasize scene contexts and spatial arrangements (e.g., ``a [subject] in a forest'', ``a [subject] on top of a purple rug''), while live prompts focus on accessories and contextual variations (e.g., ``a [subject] wearing a santa hat'', ``a [subject] in a firefighter outfit'').
Table~\ref{tab:supp_prompt_templates} shows representative examples from each category.
At generation time, each method uses the same DreamBooth prompt template and differs only in how the subject slot is filled (e.g., $S^*$ for embedding methods, \texttt{sks} for fine-tuning methods, and the class noun for \mymethod{}).
For CLIP-T evaluation, we compute text-image similarity using the class-normalized prompt for all methods, replacing method-specific identifiers with the corresponding class noun so that the metric is comparable across approaches.

\noindent\textbf{Metrics.}
We evaluate subject fidelity using DINO~\cite{dino} and CLIP-I~\cite{clip}, and editability using CLIP-T~\cite{clip}.
To isolate subject evaluation from background changes, we additionally compute Masked DINO and Masked CLIP-I using automatically extracted subject masks (detailed in Appendix~\ref{sec:supp_mask_evaluation}).
We report an overall metric as $\text{Overall} = \frac{1}{2}\!\bigl(\frac{\text{CLIP-I}+\text{DINO}}{2}+\text{CLIP-T}\bigr) = \frac{1}{4}(\text{CLIP-I} + \text{DINO}) + \frac{1}{2}\text{CLIP-T}$ , which first averages the two fidelity scores and then equally weights the combined fidelity and editability.

\begin{table}[t]
\centering
\caption{\textbf{Representative prompt templates from DreamBooth benchmark.}}
\label{tab:supp_prompt_templates}
\small
\begin{tabular}{l}
\toprule
    \textbf{Non-live Subjects (25 prompts total)} \\
\midrule
    a [subject] in the jungle \\
    a [subject] on the beach \\
    a [subject] with a tree and autumn leaves in the background \\
    a [subject] on top of a dirt road \\
    a [subject] on top of a white rug \\
    ... \\
\midrule
    \textbf{Live Subjects (25 prompts total)} \\
\midrule
    a [subject] wearing a red hat \\
    a [subject] in a firefighter outfit \\
    a [subject] wearing pink glasses \\
    a [subject] on a cobblestone street \\
    a [subject] on top of a wooden floor \\
    ... \\
\bottomrule
\end{tabular}
\end{table}

\section{Mask-based evaluation protocols}
\label{sec:supp_mask_evaluation}

\begin{figure}[t]
\centering
\includegraphics[width=1.0\linewidth]{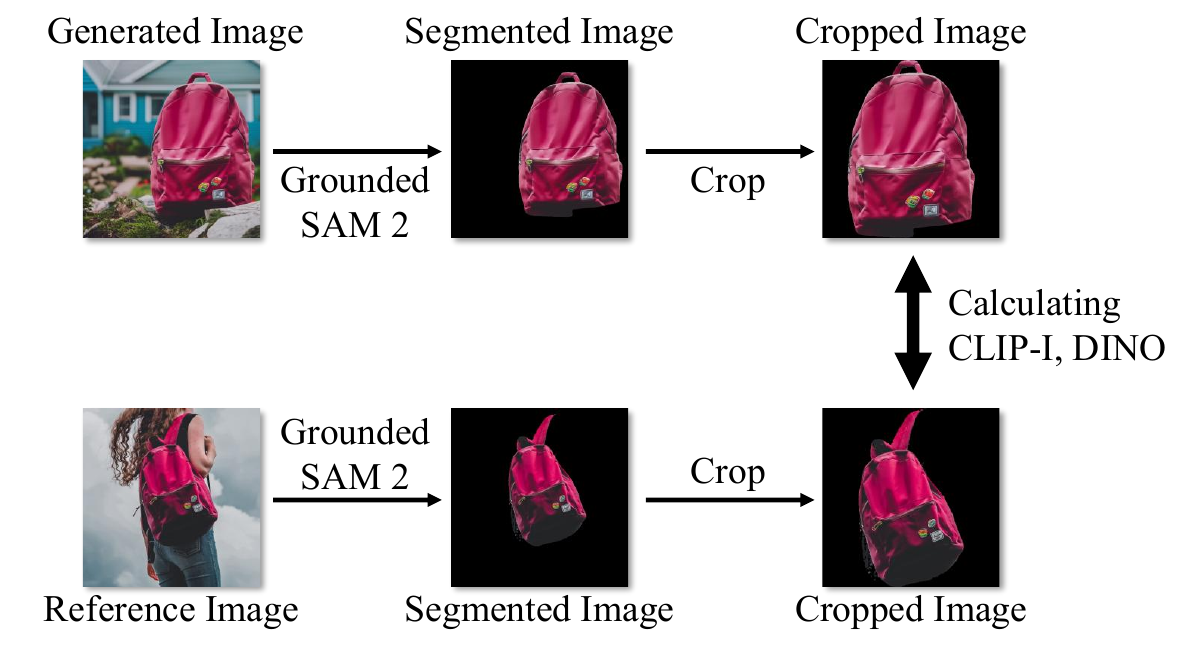}
\caption{\textbf{Overview of mask-based evaluation protocols (Masked CLIP-I \& Masked DINO).}
For both generated and reference images, we apply Grounded SAM 2 to extract subject masks, crop the regions of interest, and compute CLIP-I and DINO similarities between the cropped subjects.}
\label{fig:supp_mask_protocol}
\end{figure}

Standard fidelity metrics such as CLIP-I and DINO compute similarity over the full generated image, meaning that background similarity can inflate scores even when subject identity is poorly preserved---or, conversely, penalize a method that places the correct subject in a genuinely novel scene.
To disentangle subject fidelity from background variation, we introduce masked versions of these metrics that evaluate only the segmented subject region.
This section describes the computation protocol in detail, as illustrated in Figure~\ref{fig:supp_mask_protocol}. Note that the limitation of external segmenters discussed in Section~\ref{sec:ablation} of the main paper concerns iterative mask extraction from intermediate noisy latents during sampling; for post-hoc evaluation here, Grounded SAM 2 is applied only once to the final decoded images, where the inputs are in-distribution natural images.

To compute Masked DINO and Masked CLIP-I, we first generate masks for subject regions using Grounded SAM 2, which combines Grounding DINO~\cite{groundingdino} and SAM 2~\cite{sam2}.
We provide general class prompts (e.g., ``dog'', ``backpack'') to produce binary masks for both the generated and reference images.
After obtaining the masks, we segment the subject regions and crop them to create isolated subject images, as shown in Figure~\ref{fig:supp_mask_protocol}.
Finally, we calculate the DINO and CLIP-I similarities between these cropped subject regions.

Formally, given reference image $I_{\text{ref}}$ and generated image $I_{\text{gen}}$, we obtain their corresponding binary masks $M_{\text{ref}}$ and $M_{\text{gen}}$ through Grounded SAM 2, then crop the masked regions:
\begin{align}
I'_{\text{ref}} &= \text{Crop}(I_{\text{ref}} \odot M_{\text{ref}}), \nonumber \\
I'_{\text{gen}} &= \text{Crop}(I_{\text{gen}} \odot M_{\text{gen}}), \nonumber \\
\text{Masked DINO} &= \text{sim}_{\text{DINO}}(I'_{\text{ref}}, I'_{\text{gen}}), \nonumber \\
\text{Masked CLIP-I} &= \text{sim}_{\text{CLIP}}(I'_{\text{ref}}, I'_{\text{gen}}).
\end{align}
where $\odot$ denotes element-wise multiplication, and Crop extracts the bounding box region of the masked subject.
For the component ablation in Table~\ref{tab:component_ablation} of the main paper, the overall score is computed with masked fidelity metrics: $\text{Overall} = \frac{1}{4}(\text{MCLIP-I} + \text{MDINO}) + \frac{1}{2}\text{CLIP-T}$, because the goal of that experiment is to isolate subject-region fidelity from background variation introduced by each masking configuration.

\section{Additional quantitative results}
\label{sec:supp_extended_quant}

\begin{table*}[t]
\centering
\caption{\textbf{Additional quantitative results on SD v1.5.}
CLIP-I and DINO measure full-image subject fidelity; MCLIP-I and MDINO denote Masked CLIP-I and Masked DINO, which isolate the subject region from background; CLIP-T measures editability.
$\text{Overall} = \frac{1}{4}(\text{CLIP-I} + \text{DINO}) + \frac{1}{2}\text{CLIP-T}$.
\mymethod{} consistently improves fidelity and overall score across embedding optimization and fine-tuning paradigms on this backbone.}
\label{tab:supp_extended_quant}
\scriptsize
\setlength{\tabcolsep}{3pt}
\renewcommand{\arraystretch}{1.3}
\resizebox{0.7\textwidth}{!}{%
\begin{tabular}{@{}llcccc@{\hspace{4pt}\vrule\hspace{4pt}}cc@{}}
\toprule
\textbf{Paradigm} & \textbf{Method}
& CLIP-I $\uparrow$ & DINO $\uparrow$ & CLIP-T $\uparrow$ & \textbf{Overall} $\uparrow$
& MCLIP-I $\uparrow$ & MDINO $\uparrow$ \\
\midrule
Standalone & \mymethod{} (Ours)  & 0.7985 & 0.6724 & 0.2999 & $\mathbf{0.5177}$ & $\mathbf{0.9037}$ & $\mathbf{0.8045}$ \\
\cmidrule(l){1-8}
\multirow{6}[3]{*}{Embedding}
                                  & TI                  & 0.7755 & 0.5695 & 0.2471 & 0.4598          & 0.8470 & 0.6631 \\
 &                                \quad + \mymethod{} & 0.8303 & 0.6932 & 0.2763 & $\mathbf{0.5190}$ & $\mathbf{0.9028}$ & $\mathbf{0.8084}$ \\
\cmidrule(l){2-8}
& P+                  & 0.7595 & 0.5657 & 0.2977 & 0.4802          & 0.8626 & 0.6893 \\
 &                                \quad + \mymethod{} & 0.8041 & 0.6713 & 0.3006 & $\mathbf{0.5192}$ & $\mathbf{0.9034}$ & $\mathbf{0.8064}$ \\
\cmidrule(l){2-8}
& NeTI                & 0.7972 & 0.6334 & 0.2888 & 0.5021          & 0.8847 & 0.7533 \\
 &                                \quad + \mymethod{} & 0.8209 & 0.6954 & 0.2926 & $\mathbf{0.5254}$ & $\mathbf{0.9099}$ & $\mathbf{0.8258}$ \\
\cmidrule(l){1-8}
\multirow{4}[1]{*}{Fine-tuning}
                                  & DB                  & 0.8283 & 0.6868 & 0.2768 & 0.5172          & 0.8939 & 0.7898 \\
 &                                \quad + \mymethod{} & 0.8347 & 0.7019 & 0.2732 & $\mathbf{0.5208}$ & $\mathbf{0.8975}$ & $\mathbf{0.7976}$ \\
\cmidrule(l){2-8}
& CoRe                & 0.8280 & 0.6848 & 0.2668 & 0.5116          & 0.9010 & 0.8003 \\
 &                                \quad + \mymethod{} & 0.8387 & 0.7101 & 0.2689 & $\mathbf{0.5217}$ & $\mathbf{0.9107}$ & $\mathbf{0.8240}$ \\

\bottomrule
\end{tabular}
}
\end{table*}

Table~\ref{tab:supp_extended_quant} reports quantitative results on SD v1.5, supplementing the main paper (which covers SD v2.1, SDXL, and SD v3.5) with MCLIP-I and MDINO scores that isolate subject fidelity from background changes (evaluation protocol in Appendix~\ref{sec:supp_mask_evaluation}).
The results on SD v1.5 mirror the trends observed on SD v2.1 in the main paper.
Among embedding optimization methods, TI exhibits the largest fidelity deficit, and attaching \mymethod{} yields the most pronounced improvement: MDINO rises from 0.6631 to 0.8084, indicating that pathway separation recovers subject details that unified prompts tend to suppress.
P+ and NeTI show similar gains, with the overall score improving consistently across the three embedding baselines.
For fine-tuning methods, the improvement is more moderate---DB shows a modest gain in overall score---which is expected because fine-tuning methods already encode subject identity through adapted model weights rather than solely through attention competition.
Across both paradigms, MCLIP-I and MDINO improve in tandem with standard CLIP-I and DINO, confirming that the gains reflect genuine subject fidelity rather than incidental background matching.


\section{Additional qualitative results}
\label{sec:supp_additional_results}

\subsection{Qualitative comparison on SD v2.1}
\label{sec:supp_sd21_qualitative}

\begin{figure*}[t]
\centering
\includegraphics[width=\textwidth]{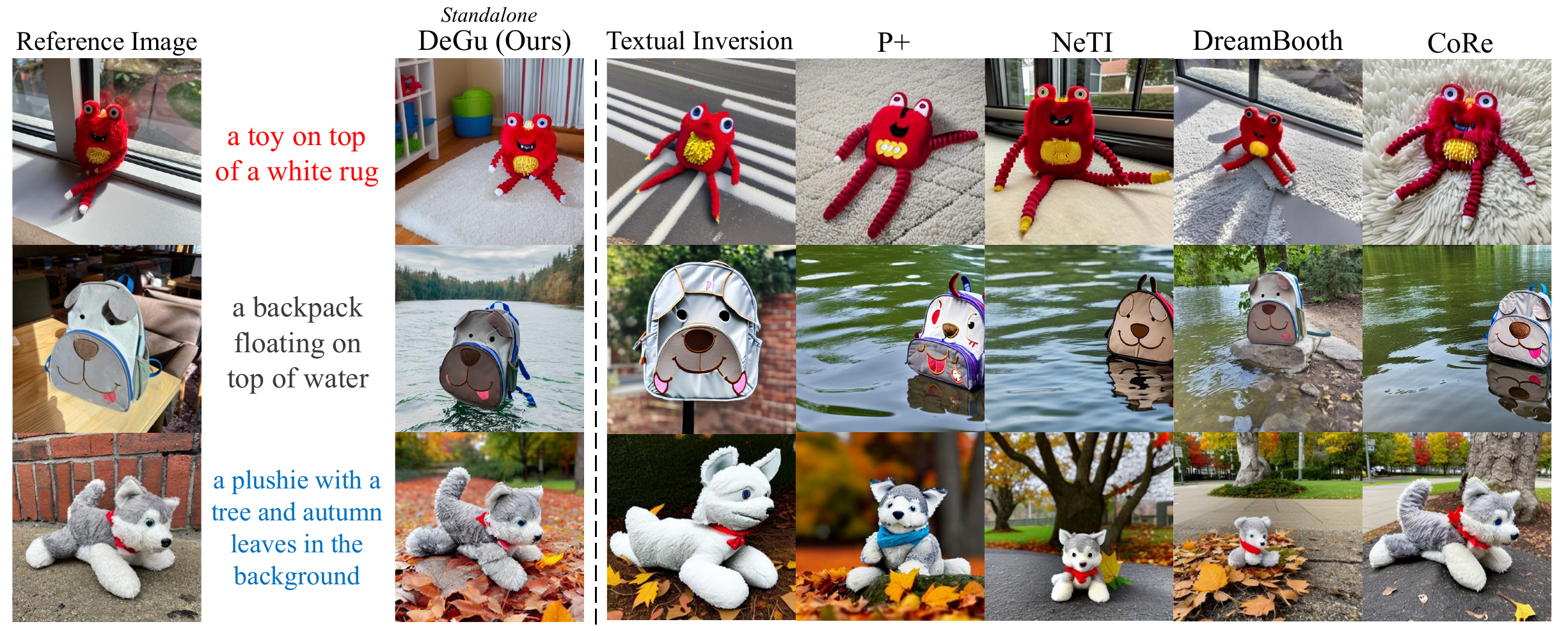}
\caption{
\textbf{Qualitative comparison on SD v2.1.}
Each row shows one subject with reference images (leftmost), \mymethod{} and outputs from five baselines.
Embedding optimization methods use ``a $S^*$ \ldots'', fine-tuning methods use ``a \texttt{sks} \ldots'', and \mymethod{} uses the plain class label (e.g., ``dog'', ``backpack'') as $\vc_e$.
Embedding optimization methods follow context prompts but lose subject identity;
fine-tuning methods preserve identity but fail to compose novel scenes.
\mymethod{} achieves both.
}
\label{fig:main_qualitative_sd21}
\end{figure*}

Figure~\ref{fig:main_qualitative_sd21} presents standalone qualitative comparisons on SD v2.1 across embedding optimization (Textual Inversion, P+, NeTI) and fine-tuning (DreamBooth, CoRe) baselines.
Embedding optimization methods follow the context prompt but lose subject-specific details such as distinctive textures and color patterns, consistent with the attention collapse analyzed in Section~\ref{sec:analysis}.
Fine-tuning methods better preserve subject identity but tend to replicate reference backgrounds rather than composing the subject into the novel scene described by the prompt.
\mymethod{} achieves both high subject fidelity and faithful scene composition by routing the two objectives through independent pathways.

\begin{figure*}
\centering
\includegraphics[width=\textwidth]{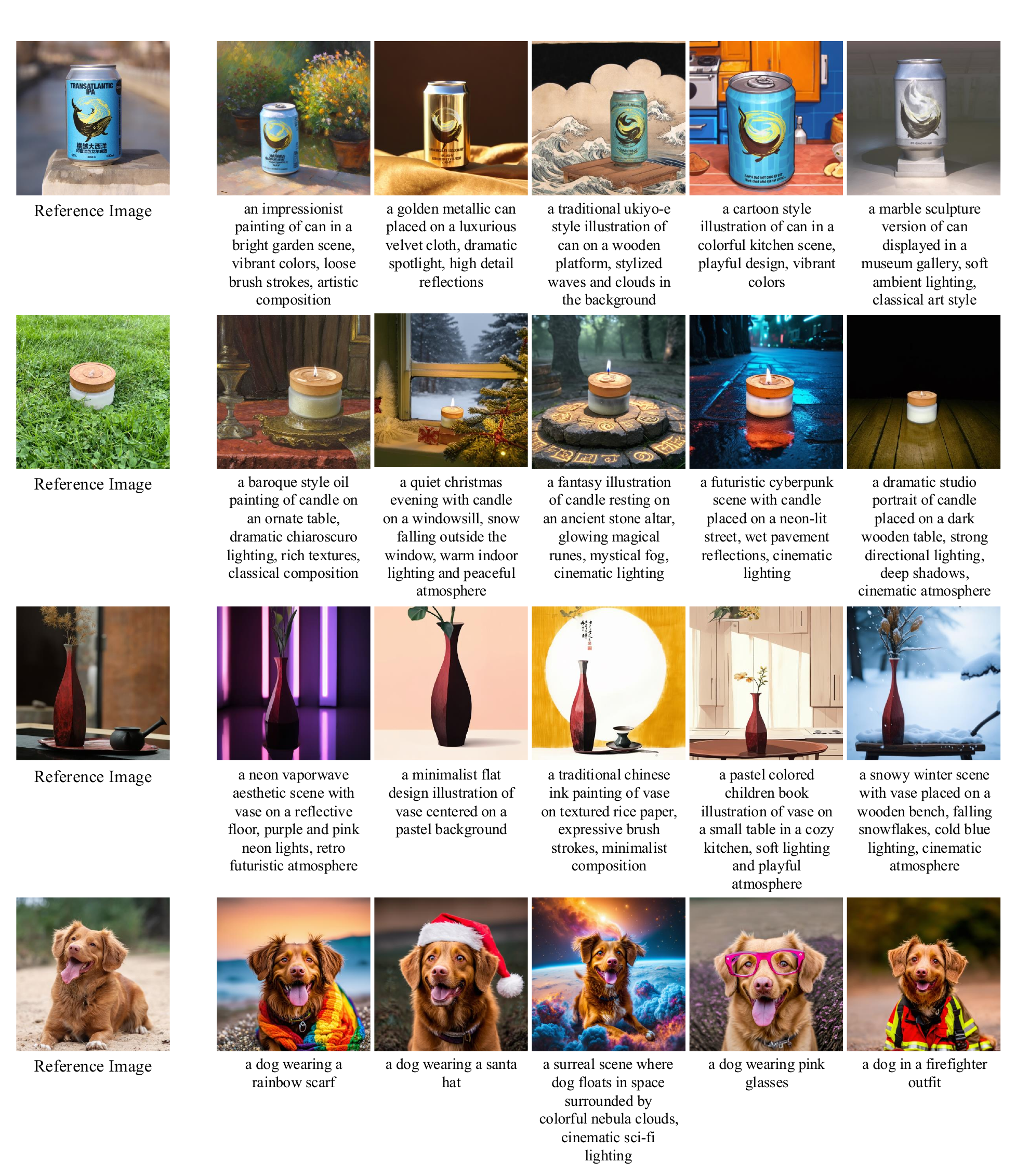}
\caption{\textbf{Qualitative results with SD v3.5 backbone.} \mymethod{} generalizes to the DiT architecture of SD v3.5, consistently preserving subject identity across diverse compositional prompts.}
\label{fig:supp_sd35_results}
\end{figure*}

\subsection{Results with SD v3.5 backbone}
\label{sec:supp_sd35_results}
To test whether the decoupled guidance mechanism transfers to architectures that differ substantially from the U-Net backbone, we apply \mymethod{} to SD v3.5, which replaces the U-Net with a Diffusion Transformer (DiT) and fuses text and image tokens through joint attention (MMDiT) rather than separate cross-attention layers.
Figure~\ref{fig:supp_sd35_results} presents standalone \mymethod{} results across four representative subjects under a wide range of compositional prompts, including style transfer (\eg, impressionist, ukiyo-e, cyberpunk), scene placement (\eg, christmas windowsill, museum gallery), and attribute binding (\eg, wearing a santa hat, floating in space).
Across these challenging conditions, \mymethod{} retains fine-grained subject details such as label typography on the can, wood-grain texture on the candle, and the curved silhouette of the vase, while faithfully reflecting the scene context specified by each prompt.
These results indicate that conditioning entanglement is not an artifact specific to the U-Net cross-attention design; it persists under joint-attention formulations as well, and pathway-level separation remains an effective remedy regardless of the underlying architecture.

\begin{figure*}[t]
\centering
\includegraphics[width=1.0\textwidth]{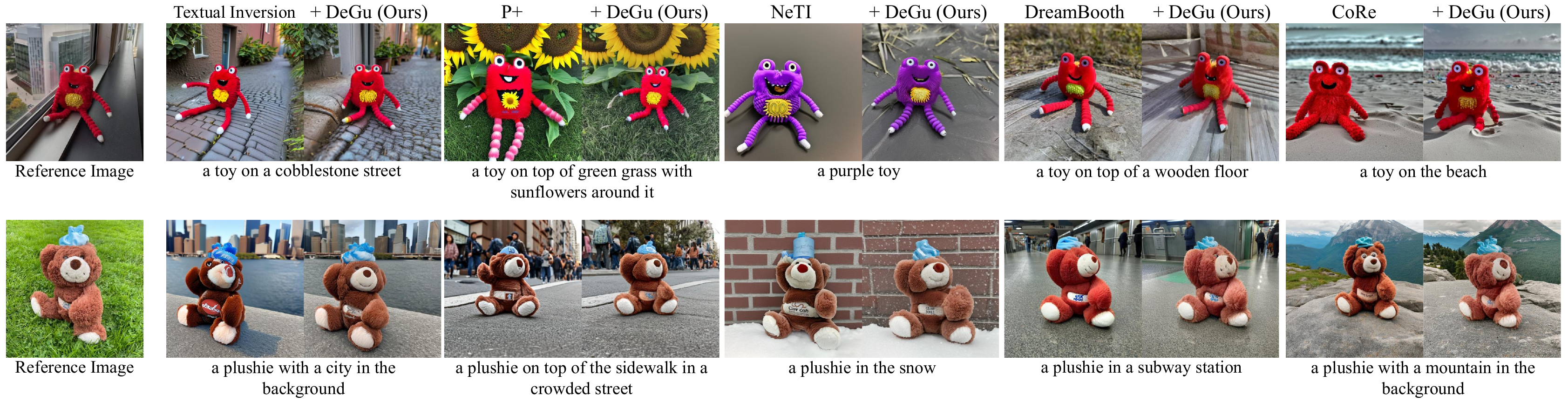}
\caption{\textbf{\textcolor{black}{Plug-and-play integration results on SD v2.1.}} 
For each baseline (TI, P+, NeTI, DreamBooth, CoRe), the left image shows the baseline output and the right shows the result after applying \mymethod{}.
\mymethod{} consistently recovers subject-specific details while preserving text-prompt adherence, using the same CAE weights trained once on SD v2.1 without any baseline-specific retraining.}
\label{fig:supp_plugandplay_results}
\end{figure*}

\subsection{Plug-and-play integration on SD v2.1}
\label{sec:supp_plugandplay}

Figure~\ref{fig:supp_plugandplay_results} shows plug-and-play integration results on SD v2.1 baselines (TI, P+, NeTI, DreamBooth, CoRe), complementing the SDXL results in the main paper.
For each baseline, the left image shows the original output and the right shows the result after attaching \mymethod{}.
Across the five baselines, a consistent pattern emerges: embedding optimization methods (TI, P+, NeTI), which tend to lose subject identity under unified prompts, recover fine-grained details such as distinctive textures and color patterns after integration.
Fine-tuning methods (DreamBooth, CoRe) retain their strong identity preservation while gaining compositional flexibility---for instance, backgrounds that previously replicated reference scenes now reflect the target prompt.
Importantly, \mymethod{} uses CAE token embeddings trained once on SD v2.1 and reuses them across the five baselines without any method-specific retraining, demonstrating the practical benefit of the plug-and-play design.


\section{User study}
\label{sec:supp_user_study}

\begin{figure}[t!]
\centering
\includegraphics[width=0.9\columnwidth]{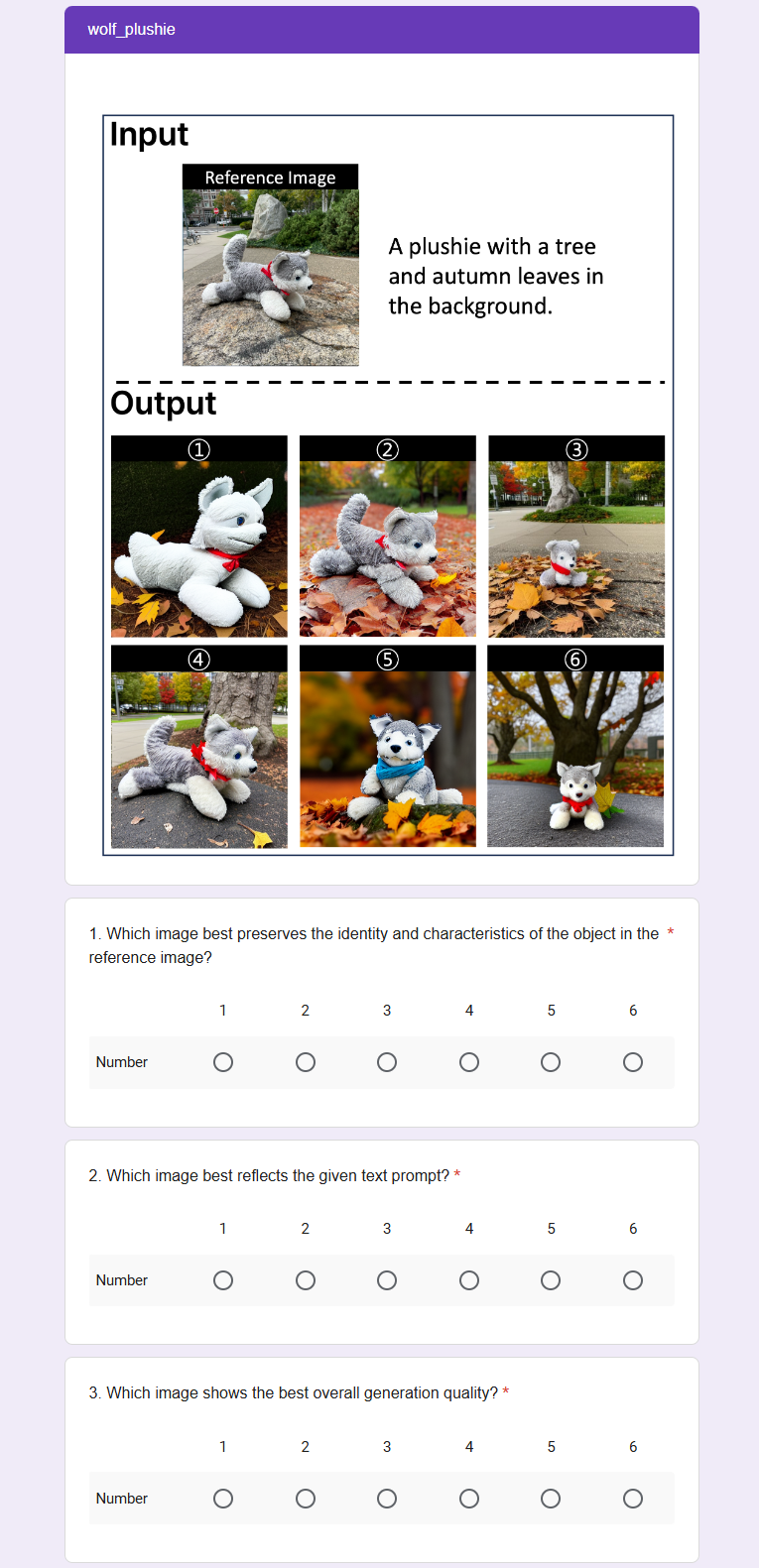}
    \caption{\textbf{User study interface.}
    For each prompt-reference pair, participants viewed reference images (top) alongside outputs from all methods, presented in randomized order.
    They then selected the best image under each evaluation criterion.
    Method names were not revealed during evaluation.}
\label{fig:supp_user_study_interface}
\end{figure}

We conducted two online survey-based user studies to evaluate the perceptual quality of generated images.
Both studies used the same evaluation protocol: participants were presented with prompt-reference pairs and, for each pair, selected the best image among the compared methods on three criteria:
\begin{itemize}
    \item ``Which image best preserves the identity and characteristics of the object in the reference image? (Fidelity)''
    \item ``Which image best reflects the given text prompt? (Editability)''
    \item ``Which image shows the best overall generation quality? (Overall)''
\end{itemize}
In both studies, the presentation order was randomized and method names were anonymized.
The two studies were conducted separately with different participant pools.
Figure~\ref{fig:supp_user_study_interface} shows the evaluation interface presented to participants.

\noindent\textbf{Results on SD v2.1.}
The first study recruited 104 participants, each evaluating 16 prompt-reference pairs across six methods (Textual Inversion, P+, NeTI, DreamBooth, CoRe, and \mymethod{}).
As shown in Table~\ref{tab:user_study}, \mymethod{} achieves the highest selection rates across the three criteria, with 49.7\% for identity preservation, 43.4\% for text prompt adherence, and 47.9\% for overall quality.
These rates are more than twice as high as those of the second-best method, CoRe (23.9\%, 20.9\%, 23.1\%), with margins exceeding 20 percentage points (each criterion aggregates $104 \times 16 = 1{,}664$ participant-pair comparisons).
Embedding optimization methods receive notably low selection rates---Textual Inversion at 1.9\% for identity and P+ at 5.2\%---which aligns with the attention collapse analysis in Section~\ref{sec:analysis}: these methods suffer the most from conditioning entanglement because their subject representations compete directly with context tokens for the same attention budget.
Fine-tuning methods (DreamBooth, CoRe) perform better but still fall substantially below \mymethod{}, suggesting that parameter-level adaptation does not fully resolve the structural interference between subject and context pathways.

\begin{table}[t!]
\centering
    \caption{\textbf{User study results on SD v2.1.} 
        For each of 16 prompt-reference pairs, participants selected the best generation among six methods across three criteria: subject identity preservation, text prompt adherence, and overall quality. 
        Values show selection rate (\%) across all responses.}
\label{tab:user_study}
\small
\setlength{\tabcolsep}{2pt}
\resizebox{1.0\columnwidth}{!}{%
\begin{tabular}{@{}lccc@{}}
    \toprule
    Method & Identity (\%) $\uparrow$ & Prompt (\%) $\uparrow$ & Overall (\%) $\uparrow$ \\
    \midrule
    Textual Inversion~\cite{textualinversion}   & 1.9  & 4.2 & 2.6 \\
    P+~\cite{p+}                                & 5.2 & 12.0 & 6.9 \\
    NeTI~\cite{neti}                            & 10.6 & 12.3 & 12.1 \\
    DreamBooth~\cite{dreambooth}                & 8.7 & 7.2 & 7.4 \\
    CoRe~\cite{core}                            & 23.9 & 20.9 & 23.1 \\
    \midrule
    \mymethod{} (standalone)            & $\mathbf{49.7}$ & $\mathbf{43.4}$ & $\mathbf{47.9}$ \\
    \bottomrule
\end{tabular}
}
\end{table}

\begin{table}[t]
\centering
\caption{\textbf{User study results on SDXL.}
Following the same protocol as Table~\ref{tab:user_study} with a separate set of participants.
Values show selection rate (\%) across all responses.}
\label{tab:user_study_sdxl}
\small
\setlength{\tabcolsep}{2pt}
\resizebox{1.0\columnwidth}{!}{%
\begin{tabular}{@{}lccc@{}}
    \toprule
    Method & Identity (\%) $\uparrow$ & Prompt (\%) $\uparrow$ & Overall (\%) $\uparrow$ \\
    \midrule
    DTI~\cite{dti}                          & 7.0 & 29.6 & 26.0 \\
    MS-Diffusion~\cite{ms-diffusion}         & 10.2 & 30.6 & 24.3 \\
    \midrule
    \mymethod{} (standalone)                 & $\mathbf{82.8}$ & $\mathbf{39.8}$ & $\mathbf{49.7}$ \\
    \bottomrule
\end{tabular}
}
\end{table}

\noindent\textbf{Results on SDXL.}
The second study recruited 31 participants, each evaluating 6 prompt-reference pairs across three methods (DTI, MS-Diffusion, and standalone \mymethod{}).
Table~\ref{tab:user_study_sdxl} reports the results.
\mymethod{} achieves the highest selection rates in identity preservation (82.8\%) and overall quality (49.7\%), while also leading in prompt adherence (39.8\%).
The identity gap is particularly pronounced: \mymethod{} surpasses MS-Diffusion (10.2\%) and DTI (7.0\%) by wide margins, indicating that participants perceive a clear advantage in subject fidelity when the two conditioning pathways are separated.
Prompt adherence scores are more evenly distributed across the three methods, which is consistent with the observation that all compared approaches retain reasonable text-following ability on SDXL; the key differentiator lies in preserving subject identity without sacrificing editability.
Combined with the SD v2.1 results in Table~\ref{tab:user_study}, these findings indicate that the perceptual gains from pathway separation transfer across different backbone architectures.

\section{Additional ablation studies}
\label{sec:supp_supple_ablation}

\begin{figure*}[ht]
\centering
\includegraphics[width=\textwidth]{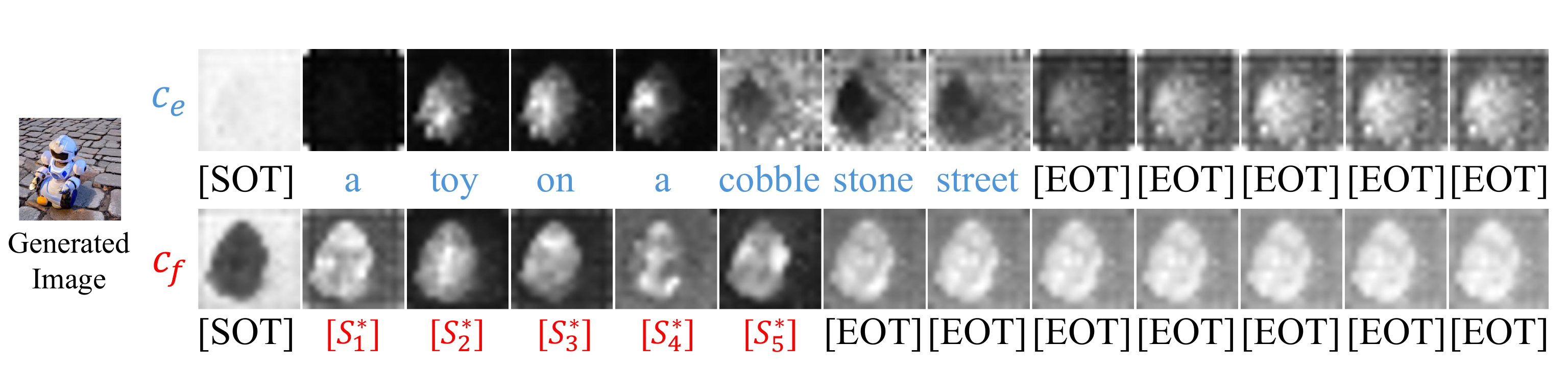}
\caption{\textbf{Token-wise attention maps in the fidelity stream $\vc_f$ for TCAM token selection (SD v2.1).} Subject tokens $[S^*]$ focus on identity-defining parts, $[\text{EOT}]$ remains diffuse, and $[\text{SOT}]$ provides the most stable complementary background response; Table~\ref{tab:supp_masking_ablation} quantifies the masking performance obtained from each candidate.}
\label{fig:supp_masking_token_comparison}
\end{figure*}

\subsection{Token capacity scaling}
\label{sec:supp_capacity_scaling_fig}
This experiment examines whether the fidelity--editability tension observed in Section~\ref{sec:analysis} of the main paper can be attributed to insufficient token capacity rather than structural attention competition.
If the trade-off were simply a matter of representational bottleneck, allocating more learnable tokens to the subject should improve fidelity without degrading editability.
Table~\ref{tab:capacity_scaling} tests this hypothesis by increasing the number of learnable tokens $n_{TI}$ in Textual Inversion from 1 to 5, with each configuration trained for 1000 steps on SD v2.1.

\begin{table}[t!]
    \centering
    \caption{\textbf{Token capacity scaling.}
    As the number of tokens for TI increases from 1 to 5, fidelity improves while editability degrades substantially.
    All models trained for 1000 steps.}
    \label{tab:capacity_scaling}
    \resizebox{1.0\columnwidth}{!}{%
    \begin{tabular}{cccc}
        \toprule
        \textbf{Token Count} & \textbf{CLIP-I} $\uparrow$ & \textbf{DINO} $\uparrow$ & \textbf{CLIP-T} $\uparrow$ \\
        \textbf{($n_{TI}$)} & \textbf{(Fidelity)} & \textbf{(Fidelity)} & \textbf{(Editability)} \\
        \midrule
        $n=1$ & 0.7625 & 0.5391 & 0.2964 \\
        $n=2$ & 0.8182 & 0.6390 & 0.2758 \\
        $n=3$ & 0.8281 & 0.6638 & 0.2676 \\
        $n=4$ & 0.8281 & 0.6691 & 0.2651 \\
        $n=5$ & 0.8290 & 0.6734 & 0.2659 \\
        \midrule
        Change ($n=1 \rightarrow 5$) & +8.7\% & +24.9\% & -10.3\% \\
        \bottomrule
    \end{tabular}%
    }
\end{table}

\begin{figure}[t]
    \centering
    \includegraphics[width=1.0\linewidth]{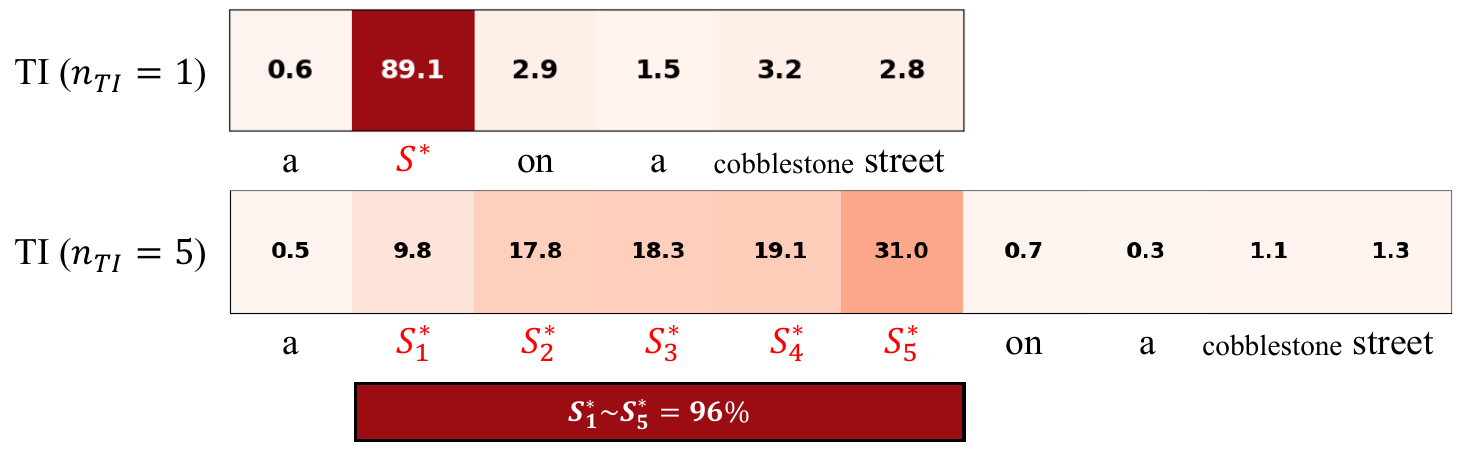}
    \caption{\textbf{Token capacity scaling intensifies attention collapse.}
    Attention heatmaps for prompt ``a \{class\} on a cobblestone street''.
    TI with a single token ($n=1$) shows severe collapse (89.1\% to $S^*$, 10.9\% to context).
    Expanding to 5 tokens ($n=5$) worsens the problem: subject tokens collectively capture 96.0\% while context tokens are suppressed to 4.0\%---a 64\% reduction.
    Despite individual token distribution (9.8\%, 17.8\%, 18.3\%, 19.1\%, 31.0\%), collective dominance intensifies, confirming that increasing capacity within unified prompts does not resolve the issue.}
    \label{fig:capacity_scaling}
\end{figure}

The quantitative trend in Table~\ref{tab:capacity_scaling} reveals a clear pattern.
Fidelity improves monotonically as the token count grows: DINO increases from 0.5391 ($n{=}1$) to 0.6734 ($n{=}5$), a gain of roughly 25\%.
However, this improvement comes at a proportional cost to editability: CLIP-T drops from 0.2964 to 0.2659, a degradation of more than 10\%.
Notably, the fidelity gains saturate beyond $n{=}3$, where CLIP-I and DINO plateau, while the editability loss continues to accumulate.
This asymmetry indicates that additional tokens are not expanding the representational capacity in a way that benefits both objectives; instead, the added tokens are absorbed into the same shared attention pool, claiming a larger collective share at the expense of context tokens.

Figure~\ref{fig:capacity_scaling} provides attention heatmap visualizations that corroborate this interpretation.
With a single token ($n{=}1$), the learned $S^*$ already captures 89.1\% of the total cross-attention, leaving only 10.9\% for context tokens such as ``cobblestone'' and ``street.''
When the token count is expanded to five ($n{=}5$), the five subject tokens $S^*_1$ through $S^*_5$ collectively absorb 96.0\% of the attention budget.
Context tokens are suppressed to 4.0\%---a 64\% relative reduction compared to the single-token case.

These results provide evidence against the capacity hypothesis and support the analysis in Section~\ref{sec:analysis} of the main paper.
The fundamental issue is not that a single token lacks representational power, but that subject tokens placed within a unified prompt compete with context tokens under the same softmax normalization.
Adding capacity inside a shared attention pool intensifies this zero-sum competition rather than relieving it, because the softmax redistributes probability mass toward whichever tokens carry stronger gradients during optimization---and subject reconstruction provides a consistently stronger learning signal than compositional context.
Decoupling the two objectives into separate conditioning pathways, as proposed in Section~\ref{sec:method}, sidesteps this competition by construction.

\begin{table*}[t]
\centering
\caption{\textbf{Effect of pathway disentanglement in plug-and-play mode (SD v2.1).}
The entangled baseline applies $\boldsymbol{c}_b$ to all conditioning terms, which reduces to two effective forward passes.
\mymethod{} uses disentangled $\boldsymbol{c}_f$ and $\boldsymbol{c}_e$ alongside $\boldsymbol{c}_b$.
The performance gap confirms that the gain originates from pathway separation rather than additional computation.
}
\label{tab:supp_disentangle}
\scriptsize
\setlength{\tabcolsep}{3pt}
\renewcommand{\arraystretch}{1.5}
\resizebox{0.9\textwidth}{!}{%
\begin{tabular}{@{}llcccc!{\hspace{4pt}\vrule\hspace{4pt}}cc@{}}
\toprule
\textbf{Method} & \textbf{Configuration}
& CLIP-I $\uparrow$ & DINO $\uparrow$ & CLIP-T $\uparrow$ & \textbf{Overall} $\uparrow$
& MCLIP-I $\uparrow$ & MDINO $\uparrow$ \\
\midrule
\multirow{2}{*}{TI}
 & Entangled ($\boldsymbol{c}_b$ only)                                              & 0.8107 & 0.6279 & 0.2748 & 0.4971 & 0.8787 & 0.7311 \\    
 & Disentangled ($\boldsymbol{c}_b$, $\boldsymbol{c}_f$, $\boldsymbol{c}_e$, Ours) & 0.8315 & 0.6888 & 0.2870 & $\textbf{0.5236}$ & \textbf{0.9045} & \textbf{0.8086} \\
\cdashline{1-8}
\multirow{2}{*}{P+}
 & Entangled ($\boldsymbol{c}_b$ only)                                              & 0.7836 & 0.6105 & 0.3075 & 0.5023 & 0.8798 & 0.7428 \\    
 & Disentangled ($\boldsymbol{c}_b$, $\boldsymbol{c}_f$, $\boldsymbol{c}_e$, Ours) & 0.8091 & 0.6719 & 0.3066 & $\textbf{0.5236}$ & \textbf{0.9037} & \textbf{0.8047} \\
\cdashline{1-8}
\multirow{2}{*}{NeTI}
 & Entangled ($\boldsymbol{c}_b$ only)                                              & 0.8107 & 0.6449 & 0.2905 & 0.5092 & 0.8891 & 0.7678 \\    
 & Disentangled ($\boldsymbol{c}_b$, $\boldsymbol{c}_f$, $\boldsymbol{c}_e$, Ours) & 0.8257 & 0.6885 & 0.2968 & $\textbf{0.5270}$ & \textbf{0.9073} & \textbf{0.8141} \\
\cdashline{1-8}
\multirow{2}{*}{DreamBooth}
 & Entangled ($\boldsymbol{c}_b$ only)                                              & 0.8032 & 0.6551 & 0.2816 & 0.5054 & 0.8868 & 0.7827 \\    
 & Disentangled ($\boldsymbol{c}_b$, $\boldsymbol{c}_f$, $\boldsymbol{c}_e$, Ours) & 0.8272 & 0.6938 & 0.2903 & $\textbf{0.5254}$ & \textbf{0.9066} & \textbf{0.8241} \\
\cdashline{1-8}
\multirow{2}{*}{CoRe}
 & Entangled ($\boldsymbol{c}_b$ only)                                              & 0.7460 & 0.5361 & 0.2939 & 0.4675 & 0.8527 & 0.6649 \\    
 & Disentangled ($\boldsymbol{c}_b$, $\boldsymbol{c}_f$, $\boldsymbol{c}_e$, Ours) & 0.7987 & 0.6514 & 0.3007 & $\textbf{0.5129}$ & \textbf{0.8974} & \textbf{0.7955} \\
\bottomrule
\end{tabular}
}
\end{table*}

\subsection{CAE token count}
\label{sec:supp_ablation_cfe_tokens}
Because CAE trains token embeddings in isolation from context tokens, an important design choice is how many tokens are needed to capture sufficient subject detail.
Too few tokens may underrepresent fine-grained identity features, while too many may increase storage cost and training time without proportional benefit.
We vary the number of context-agnostic embeddings from $n = 1$ to $7$ to determine a suitable configuration.

\begin{figure}[t]
\centering
\includegraphics[width=1.0\linewidth]{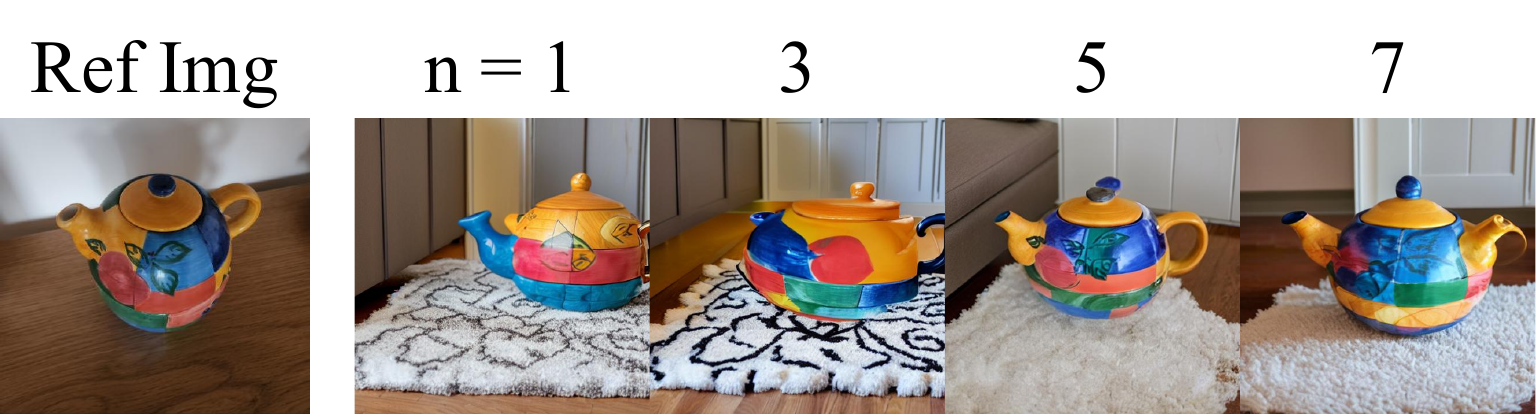}
\caption{\textbf{Effect of CAE token count on subject fidelity.} Qualitative comparison shows that $n = 5$ provides sufficient capacity to capture subject identity, with minimal improvement beyond this point.}
\label{fig:supp_cfe_token_count}
\end{figure}

Figure~\ref{fig:supp_cfe_token_count} shows qualitative results for $n = \{1, 3, 5, 7\}$ on a representative subject.
With a single token ($n{=}1$), the generated image captures the coarse category but misses distinctive details such as surface patterns and color distribution.
Increasing to $n{=}3$ recovers most of these features, and $n{=}5$ further refines fine-grained identity cues.
Beyond $n{=}5$, improvements become marginal: the $n{=}7$ output is visually indistinguishable from $n{=}5$, suggesting that five tokens provide a saturation point for the subjects in our benchmark.
We therefore select $n = 5$ as the default, which balances fidelity and storage efficiency.

\begin{figure*}[t]
\centering
\includegraphics[width=1.0\textwidth]{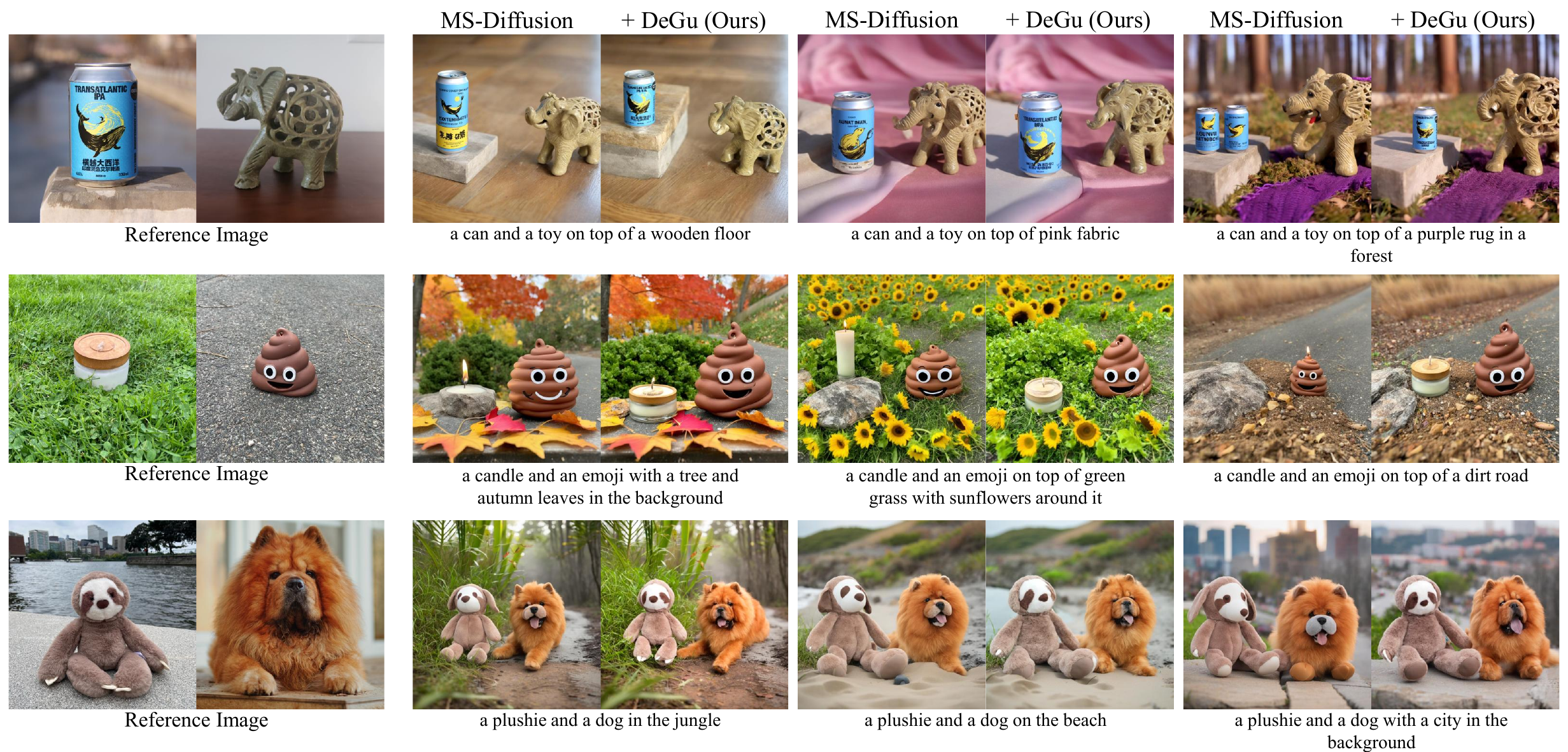}
\caption{\textbf{Multi-concept personalization.}
Each row shows two reference subjects (leftmost) and generated images from MS-Diffusion and MS-Diffusion + \mymethod{} (Ours) under the corresponding text prompt.
Attaching \mymethod{} recovers subject-specific details while preserving compositional coherence across both subjects.}
\label{fig:supp_multi_concept}
\end{figure*}

\subsection{Masking token selection}
\label{sec:supp_ablation_masking}

Because TCAM directly governs the spatial boundary between subject and background, we report MCLIP-I and MDINO here: full-image metrics would conflate subject fidelity changes with background variation introduced by the mask itself.
For spatial guidance partitioning in TCAM, the main paper establishes that non-subject tokens yield more complete subject coverage than $[S^*]$ direct attention.
Here we analyze which specific non-subject token to use in the Stable Diffusion architecture, where the $\vc_f$ sequence contains $[\text{SOT}]$, $[\text{EOT}]$, and $[S^*]$ tokens.
As shown in Figure~\ref{fig:supp_masking_token_comparison}, each token exhibits distinct attention patterns:
the subject token $[S^*]$ concentrates on identity-defining features (e.g., face, unique textures) rather than complete subject boundaries, often missing body parts;
the $[\text{EOT}]$ token shows diffuse attention extending into background regions without clear subject-background separation;
and the $[\text{SOT}]$ token naturally and consistently attends to background regions, making its inverted attention map ($1 - \mathrm{Normalize}(\mathbf{A}^{\text{SOT}}_{t,f})$) the most complete and stable subject coverage among the three candidates.

\begin{table}[t]
\centering
\caption{\textbf{Masking token ablation on SD v2.1.} SOT-based masking achieves the best subject coverage with stable mask quality.}
\label{tab:supp_masking_ablation}
\small
\begin{tabular}{lccc}
\toprule
 & MCLIP-I $\uparrow$ & MDINO $\uparrow$ & CLIP-T $\uparrow$ \\
\midrule
$[S^*]$ & 0.8993 & 0.7990 & 0.3038 \\
$[\text{EOT}]$ & 0.8938 & 0.7851 & $\mathbf{0.3043}$ \\
\textbf{$[\text{SOT}]$ (Ours)} & $\mathbf{0.9000}$ & $\mathbf{0.8019}$ & 0.3034 \\
\bottomrule
\end{tabular}
\end{table}

Table~\ref{tab:supp_masking_ablation} provides a quantitative comparison on SD v2.1.
SOT-based masking achieves the highest MCLIP-I (0.9000) and MDINO (0.8019), indicating slightly better subject coverage.
More importantly, $[\text{SOT}]$-based masks provide more consistent coverage across denoising timesteps and complex poses compared to the other two candidates. We select $[\text{SOT}]$ because TCAM prioritizes subject coverage and mask stability, and the marginal CLIP-T advantage of $[\text{EOT}]$ does not offset its weaker subject localization.
The $[S^*]$-based approach occasionally misses subject extremities (e.g., limbs, tails), while $[\text{EOT}]$-based masks tend to include background fragments that dilute the spatial partitioning.
For SDXL and SD v3.5, we use the backbone-specific non-subject token described in Section~\ref{sec:supp_train_config}.

\subsection{Disentanglement as the source of gain in plug-and-play mode}
\label{sec:supp_disentangle}

In the plug-and-play setting, \mymethod{} introduces $\boldsymbol{c}_f$ and $\boldsymbol{c}_e$ alongside the host condition $\boldsymbol{c}_b$, resulting in four forward passes per denoising step.
One might ask whether this gain could be replicated simply by applying $\boldsymbol{c}_b$ more times---that is, whether the additional forward passes alone, rather than the disentangled conditions, are responsible for the improvement.

To test this, we construct a baseline that replaces $\boldsymbol{c}_f$ and $\boldsymbol{c}_e$ with $\boldsymbol{c}_b$ while preserving the same spatial masking structure.
The mask $\mathbf{M}_t$ is derived from the cross-attention maps of $\boldsymbol{c}_b$ using the same TCAM procedure described in Section~\ref{sec:tcam}, so the comparison keeps the masking mechanism fixed and tests whether repeating an entangled condition can match the benefit of using disentangled conditions:
\begin{align}
\tilde{\boldsymbol{\epsilon}}(\mathbf{z}_t) = \boldsymbol{\epsilon}(\mathbf{z}_t, t)
&+ \gamma_b \big(\boldsymbol{\epsilon}(\mathbf{z}_t, t, \boldsymbol{c}_b) - \boldsymbol{\epsilon}(\mathbf{z}_t, t)\big) \nonumber \\
&+ \gamma_e \big[(\boldsymbol{\epsilon}(\mathbf{z}_t, t, \boldsymbol{c}_b) - \boldsymbol{\epsilon}(\mathbf{z}_t, t)) \odot (1 - \mathbf{M}_t)\big] \nonumber \\
&+ \gamma_f \big[(\boldsymbol{\epsilon}(\mathbf{z}_t, t, \boldsymbol{c}_b) - \boldsymbol{\epsilon}(\mathbf{z}_t, t)) \odot \mathbf{M}_t\big].
\label{eq:supp_entangled_baseline}
\end{align}
Because all three terms evaluate the same function $\boldsymbol{\epsilon}(\mathbf{z}_t, t, \boldsymbol{c}_b)$, Equation~\ref{eq:supp_entangled_baseline} is mathematically equivalent to applying $\boldsymbol{c}_b$ with a spatially-varying guidance scale.
This reduces to two distinct forward passes regardless of how many terms appear---meaning that repeating an entangled condition yields no additional information beyond what a single evaluation already provides.
The baseline therefore suggests that additional forward passes offer limited benefit when the condition itself remains entangled.

Table~\ref{tab:supp_disentangle} reports results on SD~v2.1 across all plug-and-play baselines.
\mymethod{} outperforms the entangled baseline across the tested methods.
Spatially distributing an entangled guidance signal does not resolve the underlying attention competition between subject and context tokens; the fidelity--editability tension documented in Section~\ref{sec:analysis} of the main paper persists because $\boldsymbol{c}_b$ continues to place both objectives within the same attention pool.
The gain of \mymethod{} therefore comes from using $\boldsymbol{c}_f$ and $\boldsymbol{c}_e$ as genuinely distinct conditions---each encoding only the semantics relevant to its target region---rather than from the number of forward passes.


\subsection{Storage-performance trade-off}
\label{sec:supp_size}

\begin{table}[t]
\centering
\caption{\textbf{Storage-performance trade-off (SD v2.1).} \mymethod{} is competitive with or exceeds the tested baselines at a fraction of the storage cost.
When combined with existing methods, it consistently improves results with negligible overhead.}
\label{tab:storage_performance}
\footnotesize
\setlength{\tabcolsep}{3pt}
\begin{tabular}{lcc!{\hspace{4pt}\vrule\hspace{4pt}}lcc}
\hline
& \multicolumn{2}{c}{\textbf{Baseline}} & & \multicolumn{2}{c}{\textbf{Combined}} \\
Method & Storage & Overall & Method & Storage & Overall \\
\hline
TI & 4KB        & 0.4953 & TI (+ \mymethod{}) & 324KB    & $\mathbf{0.5236}$ \\
P+ & 64KB       & 0.5009 & P+ (+ \mymethod{}) & 384KB    & $\mathbf{0.5236}$ \\
NeTI & 2.4MB    & 0.5077 & NeTI (+ \mymethod{}) & 2.7MB  & $\mathbf{0.5270}$ \\
DB & 3.4GB      & 0.5127 & DB (+ \mymethod{}) & 3.4GB    & $\mathbf{0.5254}$ \\
CoRe & 3.4GB    & 0.4650 & CoRe (+ \mymethod{}) & 3.4GB  & $\mathbf{0.5129}$ \\
\hline
\shortstack[l]{\rule{0pt}{2.2ex}\mymethod{}\\(standalone)\rule[-0.8ex]{0pt}{0pt}} & 320KB & $\mathbf{0.5172}$ & & & \\
\hline
\end{tabular}
\end{table}

A practical consideration for personalization methods is the per-subject storage overhead, particularly in deployment scenarios where many subjects must be stored simultaneously.
Table~\ref{tab:storage_performance} presents the storage requirements and performance on SD v2.1.

In the standalone setting, \mymethod{} achieves the highest overall performance (0.5172) while requiring only 320\,KB of storage per subject.
This is four orders of magnitude smaller than fine-tuning approaches such as DreamBooth and CoRe, which store full model checkpoints at 3.4\,GB, yet \mymethod{} achieves comparable or higher performance.
Compared to lightweight embedding methods, \mymethod{} requires more storage than TI (4\,KB) due to the use of five layer-wise CAE tokens, but the absolute overhead remains modest and the performance gain is substantial.
When integrated with existing baselines in a plug-and-play manner, the additional storage is at most 320\,KB for the CAE token embeddings---less than 0.01\% of the total checkpoint size for fine-tuning methods---making the integration cost negligible in practice.


\section{Multi-concept personalization}
\label{sec:supp_multi_concept}

The main paper evaluates single-subject personalization, where each generated image contains one target subject.
We further examine whether the decoupled guidance principle extends to multi-concept personalization, where multiple subjects appear simultaneously in a single image.
This setting intensifies the attention competition analyzed in Section~\ref{sec:analysis}, since each additional subject token further fragments the shared attention budget.
To test this, we apply \mymethod{} on top of MS-Diffusion~\cite{ms-diffusion}, which supports multi-subject zero-shot personalization via layout-guided conditioning.
As shown in Figure~\ref{fig:supp_multi_concept}, MS-Diffusion alone tends to lose fine-grained identity details of individual subjects when composing them into a shared scene, consistent with the conditioning entanglement pattern observed in the single-subject case.
Attaching \mymethod{} recovers subject-specific features for both concepts while preserving the scene context specified by the text prompt, suggesting that pathway separation remains beneficial even when the fidelity stream encodes multiple subjects at once.








\clearpage

{
    \small
    \bibliographystyle{ieeenat_fullname}
    \bibliography{main}
}

\end{document}